\documentclass[11pt]{article}

\usepackage[utf8]{inputenc}
\usepackage[english]{babel}
\usepackage[T1]{fontenc}   
\usepackage{color}
\usepackage{hyperref}

\usepackage[lofdepth,lotdepth]{subfig}
\usepackage{enumerate}
\usepackage[pdftex]{graphicx} 
\usepackage{geometry}
\definecolor{bleu}{rgb}{0, 0, 0.545098} 

\usepackage{listings}

\RequirePackage{natbib}

\usepackage{amsfonts, amsmath, amssymb}
\usepackage{amsthm}
\usepackage{mathrsfs}
\usepackage{dsfont}
\usepackage{xfrac}
\usepackage{algorithm}
\usepackage{algorithmic}
\usepackage{tikz,tkz-tab} 
\usepackage{tabularx}
\usepackage{pgfplots}

\usepackage{pifont}
\usepackage{algorithm}
\usepackage{algorithmic}
\usepackage{hyperref}
\newfloat{pseudoalgorithm}{htb}{lop}
\floatname{pseudoalgorithm}{Pseudo Algorithme}

\usepackage{tikz,tkz-tab} 
\usetikzlibrary{arrows}

\definecolor{blue1}{rgb}{0, 0, 0.5}
\definecolor{blue2}{rgb}{0, 0.3, 1}
\renewcommand{\log}{\ln}
\renewcommand{\leq}{\leqslant}
\renewcommand{\geq}{\geqslant}
\renewcommand{\phi}{\varphi}
\renewcommand{\epsilon}{\varepsilon}

\newcommand{\E}{\mathbb{E}}

\newcommand{\R}{\mathbb{R}}

\newcommand{\e}{\mathrm{e}}

\renewcommand{\H}{\mathbf{H}}
\newcommand{\M}{\mathbf{M}}
\newcommand{\W}{\mathbf{W}}
\newcommand{\p}{\mathrm{p}}

\newcommand{\transp}{\mbox{\tiny \textup{T}}}
\newcommand{\argmin}{\mathop{\mathrm{argmin}}}
\newcommand{\Var}{\mathop{\mathrm{Var}}}

\newcommand{\kl}{\mathrm{D}_{\mathrm{KL}}}

\newcommand{\titre}{Simulating Tariff Impact in Electrical Energy Consumption Profiles with Conditional Variational Autoencoders}

\title{\sf \titre}

\author{\uppercase{Margaux Br\'eg\`ere}\footnote{EDF R\&D, Palaiseau, France;
Laboratoire de math\'ematiques d'Orsay, Universit\'e Paris-Sud, CNRS, Universit\'e Paris-Saclay, Orsay, France;
INRIA - D\'epartement d'Informatique de l'\'Ecole Normale Sup\'erieure, PSL Research University, Paris, France}, AND \uppercase{Ricardo J. Bessa}\footnote{INESC Technology and Science (INESC TEC), 4200-465 Porto, Portugal}}

\begin{document}
\maketitle
\begin{abstract}
The implementation of efficient demand response (DR) programs for household electricity consumption would benefit from data-driven methods capable of simulating the impact of different tariffs schemes. This paper proposes a novel method based on conditional variational autoencoders (CVAE) to generate, from an electricity tariff profile combined with exogenous weather and calendar variables, daily consumption profiles of consumers segmented in different clusters. First, a large set of consumers is gathered into clusters according to their consumption behavior and price-responsiveness. The clustering method is based on a causality model that measures the effect of a specific tariff on the consumption level. Then, daily electrical energy consumption profiles are generated for each cluster with CVAE. This non-parametric approach is compared to a semi-parametric data generator based on generalized additive models and that uses prior knowledge of energy consumption. Experiments in a publicly available data set show that, the proposed method presents comparable performance to the semi-parametric one when it comes to generating the average value of the original data. The main contribution from this new method is the capacity to reproduce rebound and side effects in the generated consumption profiles. Indeed, the application of a special electricity tariff over a time window may also affect consumption outside this time window. Another contribution is that the clustering approach segments consumers according to their daily consumption profile and elasticity to tariff changes. These two results combined are very relevant for an \textit{ex-ante} testing of future DR policies by system operators, retailers and energy regulators. \\

\end{abstract}

\section*{Nomenclature}

\begin{table}[h!]
\centering
\begin{tabular}{clcl}
$\alpha$ & Effect of $w$ in the semi-parametric generator &   $\sigma$ & Standard  deviation of the power consumption  \\
$d$ & Dimension of CVAE latent space & $\Sigma$ & Covariance matrix \\
$E$ & Noise vector &  $\tau$ & Temperature \\
$h$ &  Half hour &   $t,\, s$ & Days   \\
$H$ & Number of half hours in a day &n$T$ & Number of days in the entire data set \\
$i$ & Household, with $\mathcal{I}$ an household set &   $T_0$ & Number of days in the training set \\
$\kappa$ & Position-in-the-year & $w$ & Type-of-day \\
 $k$ & Number of household clusters ($C_1,\dots C_k$)  & $\xi$ & Effect of $p$ in the semi-parametric generator \\
 $\mu$ & Mean of the power consumption & $x$, $X$ & Vectors of exogenous variables \\
 $N$ & Number of generated samples & $Y$ & Power consumption half-hourly profile  \\
 $p$ & Tariff, with $p \in \mathcal{P} = \{\mathrm{Low},\mathrm{Normal},\mathrm{High}\}$  &  $Z$ & Variable of CVAE latent space (decoder inputs) \\
\end{tabular}
\end{table}

\section{Introduction} 
The deployment of smart meters, which provides access to new sources of information like 5-15 minutes resolution electrical energy consumption, makes it possible to envisage the development of new customers services~\cite{mallet14}. For example, electricity demand response (DR) policies aim at modifying customers' energy consumption behavior (see \cite{SIANO2014461} for an overview) to enable higher integration levels of renewable energy sources.  \\

Most of these DR schemes rely on changes in electricity prices, which can take the form of seasonal tariffs, super-peak time-of-use, real-time pricing, critical peak pricing, etc.~\cite{Dutta2017}. Recent works (see among others~\cite{Alfaverh2020} and~\cite{bregere2019target}) proposed online learning algorithms to optimize these price incentives, considering human preferences and satisfaction level. However, the responsiveness to a tariff change may change from a consumer to another. By clustering consumers according to their tariff responsiveness, an electricity supplier can send different signals depending on the cluster to which they belong, and further improve DR management. For instance, for a given temperature, day of the week, etc., the electricity supplier defines an hourly electricity tariff profile to send to some consumers clusters. \\

A energy consumption data simulator is very useful to conduct an~\textit{ex-ante} assessment of the algorithms that set tariff profiles (i.e., ensure that they induce the right behavior from consumers) or to study the business models of different DR models~\cite{Karlsen2020} or to implement data-driven DR strategies such as contextual bandit~\cite{bregere2019target}. This simulator should be able to randomly generate energy consumption profiles for different combinations of exogenous variables and tariff profiles, with consumers clustered according to their tariff responsiveness. The present paper proposes a  novel method, based on conditional variational autoencoders (CVAE), which aims to randomly generate daily energy consumption profiles conditioned by a specific electricity tariff combined with weather and calendar variables. \\

The remainder of this paper is organized as follows. Section~\ref{sec:literature} conducts a literature review of the clustering and data generation methods applied in the energy domain and identifies the main contributions from this work. In section~\ref{sec:data}, the data set used throughout the rest of the paper is presented.
The structure of our contribution is to first provide a clustering method, in Section~\ref{sec:clustering}. Then, the CVAE approach used to generate energy consumption profiles is presented and discussed in Section~\ref{sec:CVAE}. In order to evaluate the proposed method, Section~\ref{sec:semi_param} introduces a benchmark data generator based on semi-parametric models often used for energy consumption forecasting. Section~\ref{sec:results} presents a comparison of the two generators and simulations that illustrate the interest of our approach. Section~\ref{sec:conclusions} summarizes the main conclusions and identifies potential for future work. \\

The reproducibility of this research was ensured by applying the methodology to the open data set ``SmartMeter Energy Consumption Data in London Households'' from UK Power Networks~\cite{UKdata2020}, where price incentives were sent to users via their smart meters, and by making the CVAE code available in a GitHub repository\footnote{\url{github.com/MargauxBregere/power_consumption_simulator}}.

\subsection{Motivation: generation of daily power consumption profiles for household clusters}

Since electricity is difficult to store on a large scale, its management is classically performed by anticipating demand and adjusting accordingly production.
 The deployment of smart meters, which provides access to new sources of information, makes it possible to envisage the development of new customers services (see \citealp{Yan13}; \citealp{mallet14}).
For example, electricity demand management policies aim to modify customers' energy consumption behavior, see \citet{SIANO2014461} for an overview.
This would allow to adjust to intermittency of renewable energies.
Most of them rely on changes in electricity prices.
Indeed, a higher tariff of the electricity when the electric system stability is jeopardized may induce a drop of electricity uses; and a lower tariff when electricity production is high may
encourage consumption.
The paper considers a demand response management system similar to the one experimented on some London households that took part in the UK Power Networks led Low Carbon London project in 2013\footnote{\url{https://data.london.gov.uk/dataset/smartmeter-energy-use-data-in-london-households}}: price incentives were sent to users via their smart meters. 
Recent works (see among others \citealp{oneill2010residential} and
\citealp{bregere2019target}) proposed online learning algorithms to optimize the sending of these price incentives.
The responsiveness to a tariff change may change from an consumer to another.
By considering clusters of consumers who response in the same way, electricity provider could send different signals depending on the cluster to which they belong, and further improve demand response management.
In a context similar to Low Carbon London project, for a given temperature, day of the week etc., the electricity provider defines an half-hourly electricity tariff profile to send to some costumer clusters. 
In order to test the algorithms which set tariff profiles (i.e., to ensure they make the right choices), a full-information data set is necessary: for each cluster, a realistic (but random) power consumption profile associated with each possible combination for exogenous variables and tariff profile must be generated.
We propose a method based on conditional variational auto-encoders (CVAE) to generate a random daily power consumption profile from an electricity tariff profile combined with exogenous meteorological and calendar variables. \\

In the next section, we provide a quick overview of clustering techniques and of data generation methods that we know about, in the electrical field.
At the beginning of section~\ref{sec:clustering}, we present the data set that we use throughout the rest of the paper.
The structure of our contribution is to first provide a clustering method, in Section~\ref{sec:clustering}. It relies on using a causal model which measures the effect of a tariff on half-hour power consumption.
Then, our approach is presented and discussed in Section~\ref{sec:CVAE}:
we use conditional variational autoencoders to generate power consumption profile and calibrate their hyper-parameters by grid search.
To evaluate the proposed method, we introduce, in Section~\ref{sec:semi_param}, a benchmark data generator based on semi-parametric models often used for power consumption forecasting.
Section~\ref{sec:results} concludes the paper with a comparison of the two generators and with simulations which illustrate the interest of our approach.

\section{Literature Discussion and Contributions}\label{sec:literature}

\subsection{Clustering Methods}\label{sec:clustering_rev}

Different clustering approaches were already proposed in the literature to segment consumers according to their energy consumption behavior. Generally, they relied on the construction of individual features from the average/total consumption and demographic factors. With the recent smart meter deployment, individual consumption records at higher temporal resolutions are now available and allow to consider energy consumption time series in consumers segmentation. \\

Therefore, more complex features may be extracted and used to cluster consumers with classical algorithms. Among others, Chicco et al.~compared the results obtained by using various unsupervised clustering algorithms (i.e., modified follow-the-leader, hierarchical clustering, $k$-means, fuzzy $k$-means) to group together customers with similar consumption behavior~\cite{chicco2006comparisons}; Le Ray and Pinson proposed an adaptive and recursive clustering method that creates typical load profiles updated with newly collected data~\cite{le2019online}; Rodrigues et al.~described an online hierarchical clustering algorithm, which was applied to cluster energy consumption time series in a load forecasting task~\cite{Rodrigues2008}; Fidalgo et al.~described a clustering approach based on simulated annealing that tries to reconcile billing processes that use 15 min meter data and monthly total consumption and derive typical profiles for consumers classes~\cite{Fidalgo2012}; Sun et al.~proposed a copula-based mixture model clustering algorithm that captures complex dependency structures present in energy consumption profiles and detects outliers~\cite{Sun2017}.  \\

These clustering methods do not include information about the elasticity of consumers to tariff changes. However, recent research developed mathematical and statistical models for modelling price responsiveness from domestic consumers. Ganesan et al.~applied a causality model to the Low Carbon London data set in order to rank consumers according to their responsiveness to tariff changes, and outperformed correlation-based metrics~\cite{ganesan2019using}. Saez-Gallego and Morales applied inverse optimization to improve the accuracy of load forecasting when aggregating a pool of price-responsive consumers and considering the effect of calendar and weather variables~\cite{saez2017short}; Le Ray \textit{et.al.} applied a clinical testing approach (based on a test and a control group) to assess whether or not loads of households participating in the EcoGrid EU DR program are price-responsive~\cite{le2016evaluating}; Mohajeryami et al.~proposed an economic model to explain the consumption shift between peak and off-peak hours that maximizes customer's utility function~\cite{Mohajeryami2016}.\\

These works are closely linked to the forecast of consumers reactions to DR policies, but, to our knowledge, were never combined with (or embedded in) clustering techniques for consumer segmentation or used to simulate daily consumption profiles according to their price-responsiveness.  

\subsection{Data Generation Methods}\label{sec:dgeneration_rev}

The generation of energy consumption profiles for households is not new and it was already covered by different authors in the literature. Capasso et al.~proposed a bottom-up approach based on the aggregation of individual appliance consumption in order to produce a household consumption profile~\cite{Capasso1994}. A Monte Carlo simulation model was proposed to combine behavioral data (home activities, availability at home from each member, etc.) and engineering functions (appliance mode of operation, technological penetration, etc.) with associated probability distributions. Park et al.~proposed a platform, exploiting SystemC language for event-driven simulation, which simulates the behavior of individual appliances and smart plugs~\cite{Park2010}. Both works did not considered weather-dependent appliances (e.g., heating, ventilating and air conditioning - HVAC) or the effect of price signals. \\

Physically-based models for appliances (including HVAC) are also proposed in~\cite{Muratori2013}, combined with heterogeneous Markov chain for activity patterns, to simulate households energy consumption. A similar approach was followed in~\cite{Richardson2010}, but using individual appliance consumption data. A set of physical models for appliances are proposed in~\cite{Lopez2019}, implemented in MATLAB Simulink, and can simulate optimal on/off decisions of household appliances. Gottwalt et al.~described a simulation engine for households with two modules: (a) bottom-up approach that generates consumption data for each appliance by combining statistical data about appliance use and resident presence at home; (b) optimization of appliances schedule in order to find the optimal load shift according to time-based tariffs~\cite{Gottwalt2011}. Iwafune et al.~proposed a Markov chain Monte Carlo method for simulating electric vehicle driving behaviors, which enables an evaluation of the DR potential when combined with domestic photovoltaic panels~\cite{Iwafune2020}. \\

The aforementioned methodologies assume that information about individual appliances (usage patterns, energy consumption, etc.) and behavioral data is available, instead of just using the total household consumption collected by the smart meter. One exception is~\cite{Li2019}, which describes a methodology based on an elasticity coefficient (approximated by a Gaussian distribution) to estimate indices that characterize the impact of real-time prices in the consumption pattern, such as proportion of maximum load decrease, proportion of peak-valley difference of load decrease, etc.~The method consisted in an empirical rule-based calculation of transferred consumption between periods, which was only applied to aggregated consumption of an electric power system and not to households.  

\subsection{Contributions}

The major contributions from this paper are described in the following paragraphs. \\

The CVAE-based generator of daily energy consumption profiles, in contrast to the methods revised in Section~\ref{sec:dgeneration_rev}, only relies in data collected by smart meters for the total household consumption and exogenous variables such as tariff profile, weather and calendar variables. Compared to~\cite{Park2010,Muratori2013,Richardson2010}, it is fully data-driven and does not require physical models for individual appliances and consumer behavior data. \\

Moreover, in comparison to~\cite{Li2019}, the proposed method is non-parametric and estimates changes in consumption profiles by applying a deep learning model without empirical assumptions about load shifting, showing a high capacity to learn behavioral changes when consumers experience different tariff schemes. In statistical literacy, the proposed method corresponds to sampling random vectors from a given joint density function, which was also explored in the renewable energy forecasting literature to generate temporal trajectories from conditional marginal probability distributions (see~\cite{pinson2009probabilistic} and~\cite{chen2018unsupervised} for wind energy trajectories forecast with Gaussian copula and generalized adversarial networks correspondingly). In this work, we are sampling random vectors (i.e., coherent energy consumption profiles) conditioned by tariff, weather and calendar variables. It is important to note that CVAE were recently applied in~\cite{marot2019interpreting} to learn specific representations for atypical conditions discovery (e.g., holidays) in daily electrical consumption, but not explored for synthetic data generation. \\

As a complementary contribution, a novel semi-parametric data generator, based on generalized additive models, is proposed as a benchmark model. Its numerical performance highlights the main advantage offered by the CVAE-based approach, which is the capacity to take into account and reproduce the rebound (the fall or rise in consumption shifts to another time of the day when a special tariff is applied over a period) and side (the fall or rise induced by a special tariff lasts longer -- for High tariff -- or less long -- for Low tariff -- than the period in which the tariff is actually applied) effects in the generated consumption profiles.   \\

Finally, the proposed clustering methodology extends the clustering algorithm from in~\cite{bregere2020online} in order to include the causal model between tariff and energy consumption. Thus, in contrast to the methods revised in Section~\ref{sec:clustering_rev}, this clustering algorithm gathers consumers according to their (total) consumption behavior and tariff-responsiveness.

\section{Data set Description and Preprocessing}
\label{sec:data}

As a case-study for this work, we consider the open data set published by UK Power Networks and containing energy consumption (in kWh per half-hour) of  around 5 000 households throughout 2013~\cite{UKdata2020}. A sub-group of approximately 1 100 customers was subjected to a dynamic Time of Use (ToU) tariff. The tariff values, among High (67.20 p/kWh), Low (3.99 p/kWh), or Normal (11.76 p/kWh), and the (half-hourly) intervals of day where these prices are applied, were announced day-ahead via the smart meter or text message. All non-ToU customers were on a flat rate tariff of 14.228 p/kWh and we refer to them as Standard (Std) customers. The report of Schofield \textit{et al.} (see~\cite{schofield2014residential}) provides a full description of this experimentation and an exhaustive analysis of results. \\

The data set contains tariffs and energy consumption records, for each client, at half-hourly intervals. Only ToU customers with more than 95\% of data available (1 007 clients) are kept and the same number of Std clients are sampled to build a control group. We denote by $\mathcal{I}_{ToU}$ the set of ToU households and by $\mathcal{I}_{Std}$ the set of Std ones. The missing values in the time series were filled by linear interpolation, using the previous and next interval records for small gaps and the days preceding and following for longer periods of missing data.
Finally, for each household, we also compute the average energy consumption, its minimum, and its maximum as well as the half-hour of the daily peak and of the  daily trough, for the hot months (from April to September) and for the cold months (the others).\\

Since weather has a strong impact on energy consumption, we added half-hourly data points of air temperature in London obtained from hourly public observations\cite{temp_data} by linear interpolation.
Thus, for each household $i \in \mathcal{I}_{ToU} \cup \mathcal{I}_{Std}$, for any day $t$ of year 2013, we get three $48$-vectors denoted by $Y^{1}_t(i),\dots, Y^{48}_t(i)$, $p^1_t,\dots, p^{48}_t$, and 
$\tau^1_t,\dots, \tau^{48}_t$, which are energy consumption profiles, tariff for ToU consumers and temperature respectively. From now on, $H=48$ represents the number of consumption readings per day. Since a smoothed temperature -- that models the thermal inertia of buildings -- is likely to improve forecasts (see among others,~\cite{Taylor2003} and~\cite{goude2014local}), a daily smoothed temperature $\bar{\tau}_t$ is introduced (see Appendix~\ref{app:exp_smooth} for further details). Energy consumption also depends on calendar variables such as the type-of-day and position-in-the-year. Thus, two additional variables were created: (i) binary variable $w_t$ that takes $0$ on weekends and $1$ on working days; (ii) $\kappa_t$, a continuous variable which increases linearly from $0$ (on January, 1.) to $1$ (on December, 31.).\\

The final data set (presented in Table~\ref{tab:var}) contains, for each of the 2 014 households (half Std, half ToU), $T=365$ observations of the energy consumption, tariff, and temperatures profiles, the smoothed temperature, the type-of-day, and the position-in-the-year.\\

This data set is split in two sub-sets: a training set which contains about 75\% of the original data -- days are randomly sampled from those of 2013 -- and a testing set made of the remaining data points.
A perfect design of the experiments would require four data sets but the size of the original data led us to exclude this possibility.
As the household clustering is a prior knowledge for the creation of the data generators (we create a generator per cluster), the entire data is used to cluster the clients.
The (non-parametric and semi-parametric) data generators are optimized on the training set.
The testing set is used to calibrate CVAE-based data generators and to choose the best combination of exogenous variables to give in input. 
Moreover, the best CVAE among several executions of the training process  (CVAEs may converges to local minima) is selected thanks to this testing set.
Finally, it also permits to compare the two approaches, non-parametric and semi-parametric, in the experiments of Section~\ref{sec:results}.
To simplify notation, we re-indent the observations of the original data set: observations from $1$ to $T_0=273$ form the training set, and the ones from $T_0+1$ to $T=365$ form the testing set. The dataset division and use is summarized in Table~\ref{tab:data_set}.
 
\begin{table}[h!]
\centering
\begin{tabular}{lcc}
  \hline
  & Training Set & Testing Set \\
  \hline
   \hline 
Households clustering & \checkmark & \checkmark \\
Semi-parametric model training & \checkmark & \\ 
CVAE  model training  & \checkmark & \\ 
CVAE hyper-parameters calibration & & \checkmark \\
CVAE model selection & & \checkmark \\
Numerical experiments & & \checkmark \\
  \hline
\end{tabular}
\caption{Summary of the use of the two data sets: the training set (75\% of the original data) and the testing set (remaining data). The clustering of the households is detailed in Section~\ref{sec:clustering}. The training process for the CVAE-based generator is explained in Section~\ref{sec:CVAE}; 
the calibration of the hyper-parameters and the selection of the best CVAE are detailed in the subsections~\ref{sec:hyper_cal} and~\ref{subsec:selec}, respectively.
The training process for the semi-parametric generator is in Section~\ref{sec:semi_param}. Both data generators are compared in the experiments of Section~\ref{sec:results}.}
\label{tab:data_set}
\end{table}

\begin{table*}[tb]
\centering
\begin{tabular}{llc}
  \hline
  Variable & Description & Notation\\
  \hline
   \hline 
Energy consumption & Daily energy consumption profile (half-hourly intervals) & $Y^{1}_t(i),\dots, Y^{H}_t(i)$  \\
Tariff (ToU consumers) & Daily electricity price profile (half-hourly intervals) & $p^1_t,\dots ,p^{H}_t$ \\
Temperature & Daily London air temperature profile (half-hourly intervals)  & $\tau^1_t,\dots ,\tau^{H}_t$ \\
Smooth temperature & Computed from past temperatures  & $\bar{\tau}_t$ \\
Type-of-day &  $1$ from Monday to Friday, $0$ for week-ends &  $w_t$ \\
 Position-in-the-year & Linear value between $0$ (January, 1.) and $1$ (December, 31.) & $\kappa_t$ \\
  \hline
\end{tabular}
\caption{Summary of the variables provided and created for each household $i$ of the data set.}
\label{tab:var}
\end{table*}

\section{Clustering of Household Consumers}
\label{sec:clustering}

\subsection{Causality model}\label{sec:causality}

To measure the impact of the tariff on the energy consumption, a causality model similar to the one proposed by Ganesan \textit{et al.} (see~\cite{ganesan2019using}) is considered.
The finite set of available tariff is denoted by $\mathcal{P}=\{\mathrm{Low}, \, \mathrm{High},\, \mathrm{Normal}\}$ and  its cardinal by $|\mathcal{P}|$.
For each household and each tariff, a daily profile of the mean and the standard deviation of its energy consumption will be computed. For an household $i$, at an half-hour $h$, the random variable $Y^{h}(i)$ refers to the individual energy consumption of household $i$. It depends on the chosen tariff $p \in \mathcal{P}$ but also on the exogenous variables gathered in a vector $x^h=(\tau_t^h,\bar{\tau}_t,w_t,\kappa_t)$. \\

Here, the aim is to estimate, for each tariff $p$ and for each half-hour $h$, the expectation and the standard deviation of the random variable $Y^{h}(i)\, | \, P=p$. Thanks to $T$ observations $Y^{h}_t(i)$, $x^h_t$, and $p^h_t$,  with $t \in \{1,\dots,T\}$, of energy consumption, tariffs, and exogenous variables, respectively, a model that gives, for the tariff $p$ and the exogenous variables $x^h$, a forecast of the expected consumption at $h$ when tariff $p$ is picked, is trained. In the original model, the authors used kernel regression and then an approach based on  bootstrapping to provide an estimation of the standard deviation (see~\cite{ganesan2019using} for further details). In this work, for any exogenous variable $x^h_t$ and tariff $p^h_t$, the random energy consumption $Y^{h}_t(i)$ is assumed to be Gaussian of mean $\mu_{i}(x^h_t,p^h_t)$ and standard deviation $\sigma_{i}(x^h_t,p^h_t)$ and that theses mean and standard deviation depend on additive smooth predictors. They are estimated with generalized additive models (GAM), see~\cite{wood2006generalized} -- full calculations are detailed in Appendix~\ref{app:causality_mod}. Therefore, for any tariff $p$, the trained model provides these estimations, that are denoted by $\widehat{\mu}_{i}(x^h_t,p)$ and $\widehat{\sigma}_{i}(p,x_t^h)$.
Then, an approximation of the impact of a tariff change is computed with the two following quantities:
\begin{gather}
\E\big[\,Y^{h}(i)\, | \, P=p\, \big] \approx \frac{1}{T}\sum_{t=1}^T \widehat{\mu}_{i}\big(x_t ^h,p\big) 
\quad 
\mathrm{and} \quad \sqrt{\Var\big[\,Y^h(i)\, | \, P=p\, \big]}  \approx \frac{1}{T}\sum_{t=1}^T \widehat{\sigma}_{i}\big(x_t^h,p\big)\,. \label{eq:var}
\end{gather}
For simplicity of notation, these approximations associated with an household $i \in \mathcal{I}_{ToU} \cup \mathcal{I}_{Std}$, are denoted by $\mu^{h}_i(p)$ and $\sigma^{h}_i(p)$, respectively.
Vectors  $\mu^{1}_i(p),\dots,\mu^{H}_i(p)$ will be used to cluster the consumers whereas vectors $\sigma^{1}_i(p),\dots,\sigma^{H}_i(p)$ will not be used until later, in Section~\ref{sec:semi_param} for the creation of the benchmark data generator. Actually, they will not be directly useful, but a similar approach will be applied to compute the standard deviation per tariff of the energy consumption of a consumer cluster, namely by replacing household $i$ by a group of households.

\subsection{Clustering Method} 

The proposed method used to cluster the households according to their consumption profile is very similar to the one used in~\cite{bregere2020online}.
In this section, $\mathcal{I}$ will refer indifferently to $\mathcal{I}_{ToU}$ or to $\mathcal{I}_{Std}$. 
For any household, $i \in \mathcal{I}$, the causality model described in the previous section provides, for each tariff $p \in \mathcal{P}$, a daily energy consumption profile, namely $H$ mean energy consumption $\mu^1_i(p), \dots, \mu^H_i(p)$. 
As the focus is more on the shape of the profiles, rather than on the amount of consumed electricity, the profiles of an household $i$ are first rescaled with its average consumption associated with a base tariff, namely Normal tariff.\\

Then, these profiles are concatenated in a matrix $\M \in \mathcal{M}_{|\mathcal{I}|\times H |\mathcal{P}|}$ that gathers all the households.
The dimension of $\M$ is reduced with a non-negative matrix factorization (NMF): with $r$ a small integer, $\M$ is approximated by $\W \, \H$, where $\W$  and  $\H$ are $|\mathcal{I}| \times r$ and $r \times  H|\mathcal{P}|$-non-negative matrices, respectively. 
As soon as this approximation is good enough, line $i$ of the matrix $\W$ is sufficient to reconstruct household $i$ profiles (with the knowledge of matrix $\H$ - which is not used for the clustering). 
Thus, for each household $i$, from the $H|\mathcal{P}|$-vector $( \mu^1_i(p), \dots, \mu^H_i(p))_{p \in \mathcal{P}}$, $r$ features are extracted: line $i$ of $\W$.
With this low dimension representation of households in $\mathbb{R}^r$, $k$-medoids clustering algorithm provides the $k$ clusters $C_1,\dots,C_k$,
using \texttt{KMedoid} function
 implemented in the  \texttt{Python}-library \texttt{sklearn\_extra}.
The diagram in Figure~\ref{fig:clustering} sums up the steps of the procedure described here in a summarized way and detailed in Appendix~\ref{app:clustering}.
\begin{figure}[t]
\centering
\includegraphics[width=0.95\textwidth]{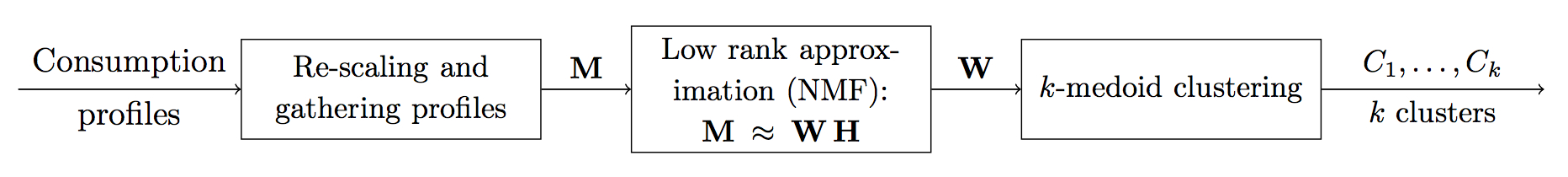} 
\caption{Scheme of the proposed clustering method.}
\label{fig:clustering}
\end{figure}

\subsection{Evaluation of the Households Clustering} 

Three different clustering approaches of the households of $\mathcal{I}_{ToU}$ and of $\mathcal{I}_{Std}$, with $k=4$ clusters, are compared. The first one is a random clustering:  an integer between $0$ and $k-1$ is randomly assigned to each household. The second one relies on classical features used to define an households profile: the minimum, maximum, and average consumption in winter and in summer, the peek-hour, and the off-hour (average instant of maximum and minimum consumption). From these rescaled features, $k$-medoid algorithm is used to cluster the households. The third approach is the one proposed in this paper and described in the previous section. For a cluster $C_{\ell}$, and for any day $t$ and half-hour $h$, we will, from now on, consider the average energy consumption $Y_t^h(C_\ell)~=~1 / |C_\ell| \sum_{i \in C_\ell} Y_t^h(i)$, where $Y_t^h(i)$ is  the energy consumption record associated with household $i$. \\

\begin{figure*}[t]
\centering
\includegraphics[width=0.4\textwidth]{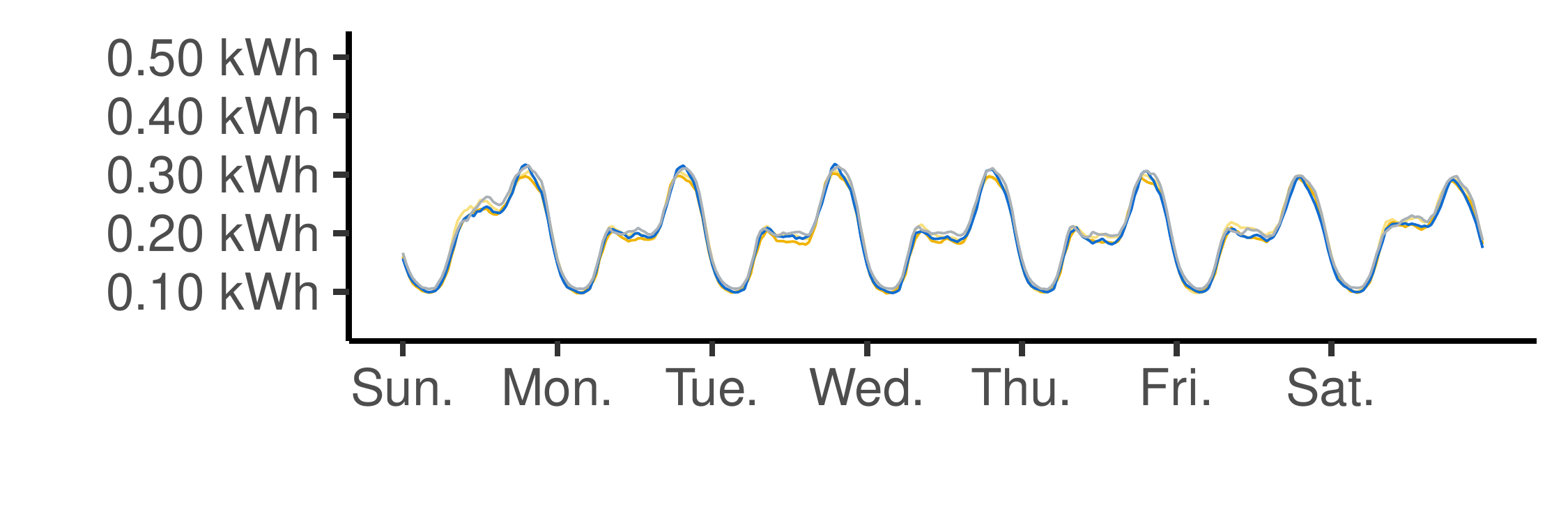} \hfill
\includegraphics[width=0.59\textwidth]{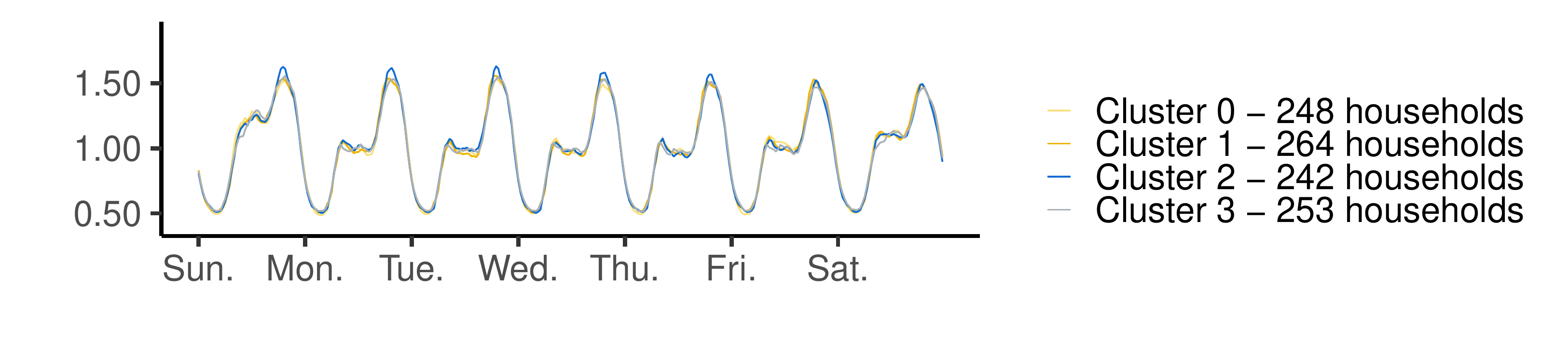} 
\includegraphics[width=0.4\textwidth]{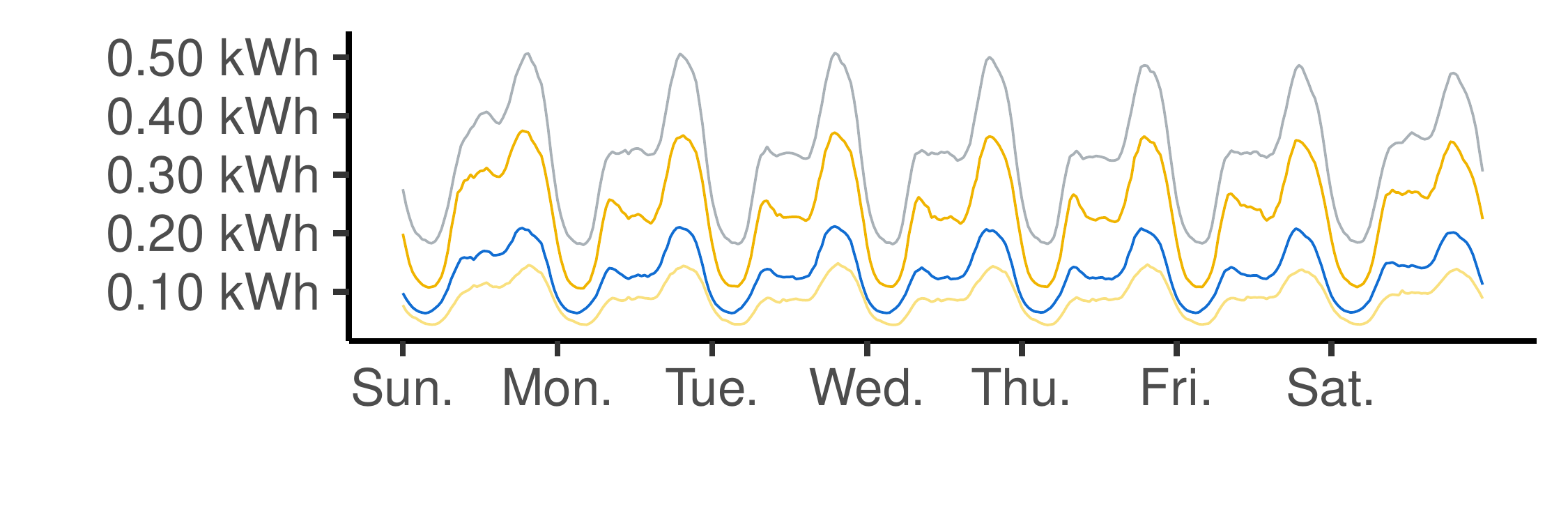} \hfill
\includegraphics[width=0.59\textwidth]{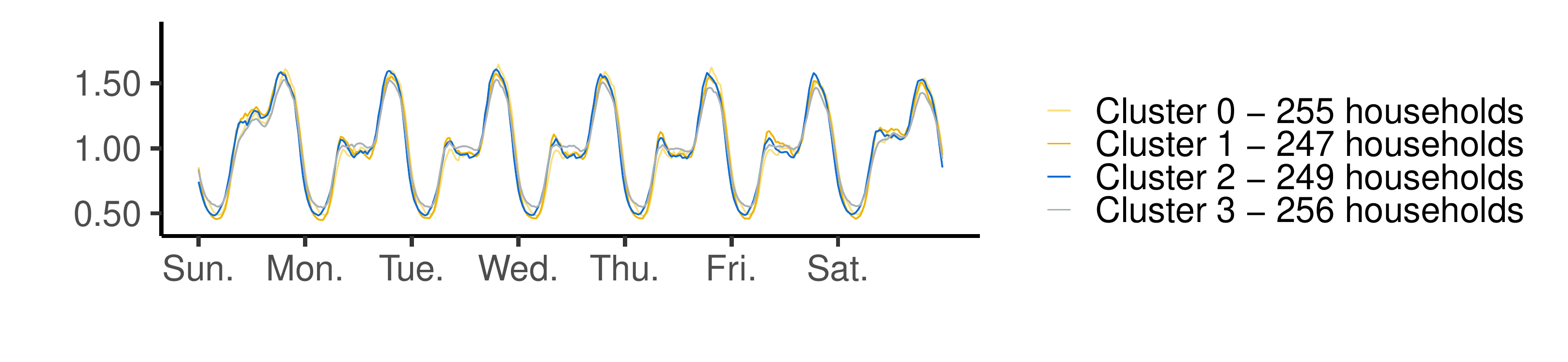} 
\includegraphics[width=0.4\textwidth]{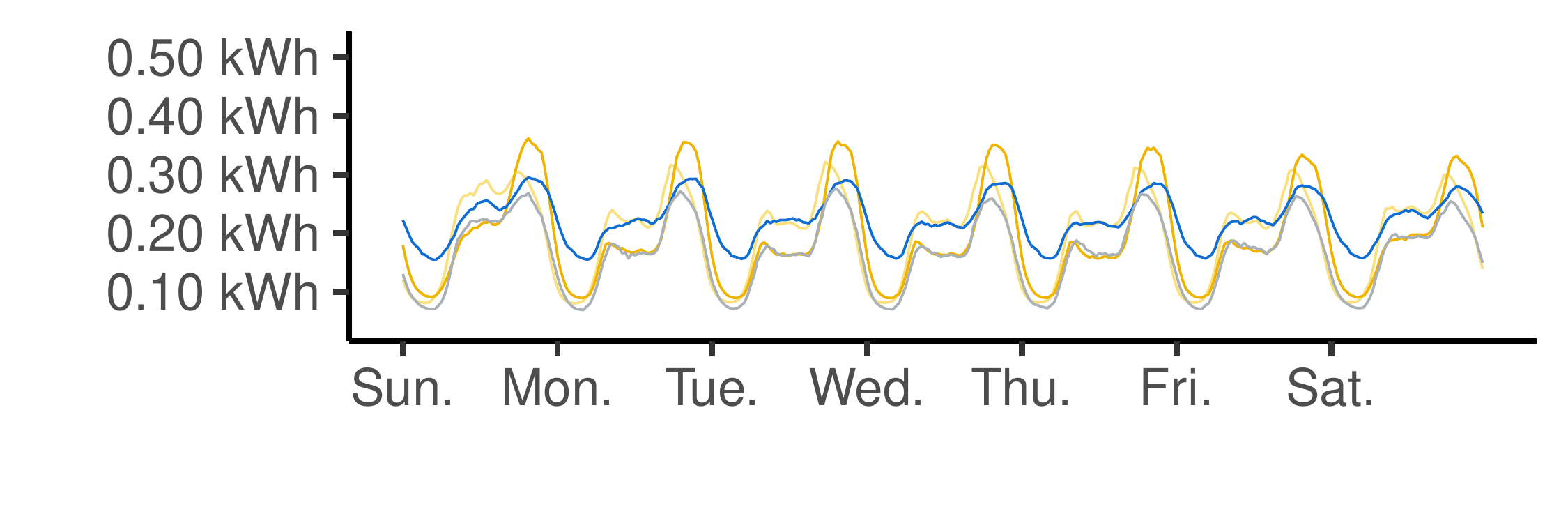} \hfill
\includegraphics[width=0.59\textwidth]{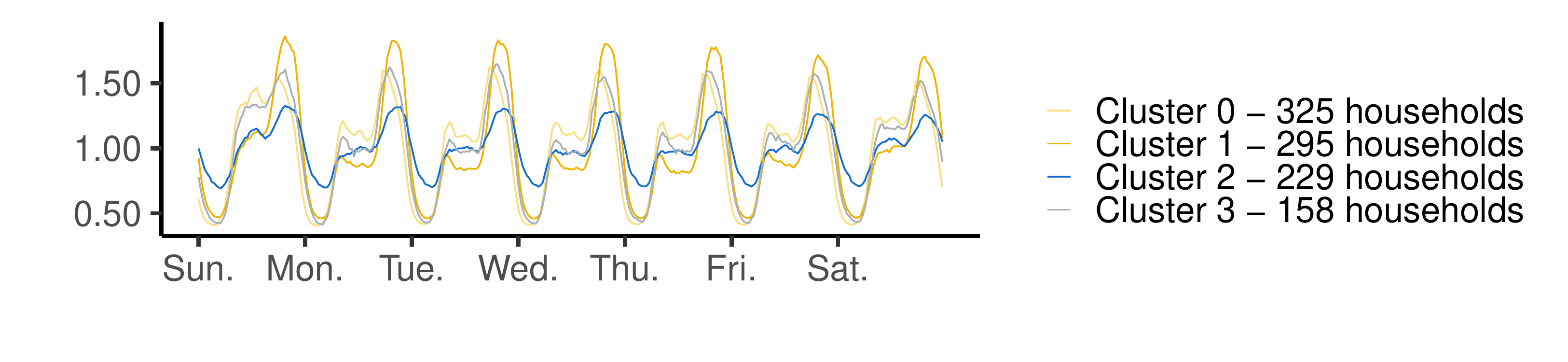} 
\caption{Daily profile of ToU cluster energy consumption (left) and standardized energy consumption (right) for a random clustering (top), for a clustering based on ``classical features'' (middle), and the clustering method proposed in Section~\ref{sec:clustering} (bottom).}
\label{fig:weekly_profiles}
\end{figure*}

Figure~\ref{fig:weekly_profiles} depicts, for the three clustering approaches applied on ToU households, the weekly profile of the average energy consumption of each cluster $Y_t^h(C_\ell)$ (to the left) and the normalized energy consumption (to the right), namely the weekly profile of $Y_t^h(C_\ell)/\big( \frac{1} {TH}\sum_{s=1}^T\sum_{j=1}^H Y_s^j(C_\ell)\big)$. 
Classical features allow to discriminate households depending on the amount of electricity they consume but does not really catch daily or weekly behavior. 
Conversely, profile types clearly come off with the proposed method.\\

\begin{table*}[t]
\centering
\begin{tabular}{lrrrrrr}
  \hline
            & Rd ToU & Features  ToU &  NMF ToU & Rd Std & Features Std &  NMF Std \\
  \hline
    \hline
Non-standardized & 4 627 &  40 764 &  13 432 & 5 014 &   5 088 & 12 971 \\
Standardized   & 4 950 & 6 870 & 14 088 &
 4 834 &   4 741 & 14 934 \\
Special tariff - standardized & 5 151 & 7 033  &16 070 & 4 904 &   4 694 & 15 460 \\
  \hline
\end{tabular}
\caption{Calinski-Harabasz score for a random clustering (``Rd''), for a clustering based on classical features (``Features''), and the clustering method proposed in Section~\ref{sec:clustering} (``NMF'') computed for different consumption record series.}
\label{tab:CHscore}
\end{table*}

The Calinski-Harabasz index, (see~\cite{calinski1974dendrite}) is a variance ratio criterion,  that evaluate the relevance of the clustering.
By denoting $Y(i)$ the vector that contains some of the consumption records associated with household $i$, and by $Y(C_\ell)$ the one with the average consumption records of cluster $C_\ell$ and by $Y (\mathcal{I})$ the average consumption records of all households,
the score $S_{\mathrm{CH}}$ is defined as the ratio of inter-clusters variances and intra-cluster variances:
\begin{gather}
S_{\mathrm{CH}} = \frac{\big(|\mathcal{I}| -k\big)\Var(C_1,\dots,C_k)}{\big(k-1\big) \sum_{\ell=1}^k\Var(C_\ell) } \\
 \mathrm{with} \quad \Var(C_1,\dots,C_k) = \sum_{\ell=1}^K \big\| Y (C_\ell)- Y (\mathcal{I})\big\|^2 
  \mathrm{and} \quad  \Var(C_\ell) = \frac{1}{|C_\ell |} \sum_{i\in C_\ell} \big\| Y(i)- Y(C_\ell) \big\|^2  \label{eq:var_inter} \,.
\end{gather}
where $\Var(C_\ell)$ is the intra-cluster variance of $C_\ell$ and $\Var(C_1,\dots,C_k)$ is the inter-clusters variance.\\

To compute this score, three different vectors $Y$ are considered.
First, all the records of the data set are taking into account, namely the records of the entire year 2013; therefore, in Equation~~\eqref{eq:var_inter}, the vector $Y(i)$  is equal to $\big(Y_1^1(i), Y_1^2(i),\dots, Y_T^H(i)\big)$.
Then we look at the normalized energy consumption records,
so $Y(i) = \big(Y_1^1(i), Y_1^2(i),\dots, Y_T^H(i)\big) / \big(\frac{1}{TH} \sum_{t=1}^T \sum_{h=1}^H Y_t^h(i)\big)$.
Finally the normalized records associated with the sending of incentive signals are selected: only the normalized records associated with tariff Low or High are kept and the others are removed.
The results are presented in Table~\ref{tab:CHscore}, where we observe a higher score on non-standardized records for the ``classical features'' clustering, which is totally coherent with the curves of Figure~\ref{fig:weekly_profiles}.
The proposed clustering method seems efficient for catching households behavior. Indeed it gets the higher score for standardized records.
Moreover, the score is even higher when we select only records associated with special tariff and this increase is more important for ToU consumers that for Std ones. This presumes that the clustering is not only catching a global behavior but also the reaction to a tariff change. \\

It is important to mention that since we want to simulate energy consumption of quite large sub-groups of households (between one and five hundreds households), we did not investigated the optimal number of clusters $k$ (i.e., it was fixed to $4$). \\

In the following sections we present the two data-driven methods that simulate energy consumption profiles associated with the clusters of $\mathcal{I}_{ToU}$ obtained with the method described above. 
For both approaches, we will train a data generator per cluster.
So from now on and for simplicity of notation, a record $Y_t^h$ will refer to $Y_t^h(C_{\ell})$, where $C_{\ell}$ designs any clusters of set $\mathcal{I}_{ToU}$.

\section{Energy consumption profile generation with conditional variational autoencoder} \label{sec:CVAE}

The training set made of the $T_0$ observations $(Y_1, X_1)$, $(Y_2,X_2), \dots $, $(Y_{T_0}, X_{T_0})$ is considered. For a day $t$, $Y_t = (Y_t^1,\dots,Y_t^H)$ is the $H$-dimension vector which corresponds to the daily profile of the half-hour energy consumption of a household cluster.  The vector $X_t$ gathers  calendar, weather, and tariff information of day $t$, which will be detailed further.

\subsection{Conditional variational autoencoder}

\subsubsection{Description}

The proposed method to generate energy consumption profiles uses the conditional version of variational autoencoders (VAE), which are generative models introduced by Kingma and Welling in 2013 (see~\cite{kingma2013auto} for further details). Autoencoders were mostly used for dimensionality reduction or feature learning (see, among others \cite{rumelhart1986learning} and \cite{ hinton1994autoencoders}).
They consist of two neural networks: an ``encoder'' $\mathrm{E}$ and a ``decoder'' $\mathrm{D}$.
An autoencoder learns a low dimension representation of a set of $H$-dimension data points by training both networks at the same time. 
Indeed, the encoder transforms the $H$-dimension vectors into $d$-dimension vectors (with $d \ll H$) and the decoder tries to rebuild initial vectors from the encoder outputs.
Considering $Z=\mathrm{E}(Y)$ as the $d$-dimension output of the encoder for the $H$-dimensional input $Y$ and $\mathrm{D}(Z)$ as the $H$-dimension output of the decoder for the $d$-dimension input $Z$, the autoencoder is trained 
to minimize the following ``reconstruction loss''
\begin{align}
L_{\mathrm{AE}} = \frac{1}{T_0}\sum_{t=1}^{T_0} \, \, \big\| Y_t-\widehat{Y}_t \big\|^2 \notag =  \frac{1}{T_0}\sum_{t=1}^{T_0} \, \, \big\| Y_t - \mathrm{D}\big(  \mathrm{E} (Y_t) \big) \big\|^2 \,,
\end{align}
where $\| \cdot \|$ is the Euclidean norm. Therefore, a data point $Y$ can be represented in a $d$-dimension latent space by $\mathrm{E}(Y)$. \\

In the autoencoder framework, there is no constraint on this latent space and the only guarantee is that the representation $Z = \mathrm{E}(Y)$ can be decoded in the original signal $\mathrm{D}(Z) \approx Y$.
Moreover, we have no idea what the decoded variable $\mathrm{D}(Z)$  would look like for a value of $Z \notin \{\mathrm{E}(Y_1),\dots,\mathrm{E}(Y_{T_0} ) \}$. 
Thus, there is no guarantee on the shape of the latent space. 
Without regularization term, for any $d\geq 1$, by increasing the number of neurons in both the encoder and the decoder networks, we can create an autoencoder with enough degrees of freedom to fully overfit the data, which 
points out the need for a regularization term. 
In VAEs, the introduction of a penalty on the latent space implicitly makes the strong assumption that the distribution of data points $E(Y)$ is close to a given prior distribution.
This prior is often set to the standard normal distribution, which we also do in our experiments.
From now on, the encoder encodes the distribution of $Z|Y$, which is wanted close to $\mathcal{N} (0,I_d)$.
We consider that $Z \,| \, Y\sim\mathcal{N}(\mu(Y),\Sigma(Y))$, where $\mu(Y)$ and $\Sigma(Y)$ are the encoder outputs.
The outputs $\widehat{Y}_t$ of the decoder are now $\mathrm{D}(Z_t)$, where the random variable $Z_t$ is sampled from a $d$-multivariate Gaussian of mean $\mu(Y_t)$ and covariance matrix $\Sigma(Y_t)$, which are the encoder outputs.
With $\kl ( P \, || \, Q )$ as the Kullback-Leibler divergence from $Q$ to $P$, the VAE is trained by minimizing the following loss 
\begin{align}
L_{\mathrm{VAE}}(\eta) =  \frac{1}{T_0}\sum_{t=1}^{T_0}  \, \, \big\| Y_t-\widehat{Y}_t \big\|^2   \quad + \,  \eta \, \frac{1}{T_0}\sum_{t=1}^{T_0}   \kl \Big( \mathcal{N}\big(\mu(Y_t),\Sigma(Y_t)\big) \,  \big|\big|  \, \mathcal{N}(0,I_{d})  \bigg) \,.
\label{eq:loss}
\end{align}
The first term corresponds to the reconstruction error and the second one is a regularization penalty on the latent space.
The coefficient $\eta$ balances these two terms. 
Calculations from~\cite{kingma2013auto} are recalled in Appendix~\ref{app:vae}. They show how, under some assumptions on the existence of a representation of the data in a $d$-dimensional latent space, minimizing this loss corresponds 
to conjointly maximizing the likelihood of the observations with the density induced by the data generation process and minimizing an approximation error in the latent space. \\

\begin{figure}[t]
\centering
\includegraphics[width=0.65\textwidth]{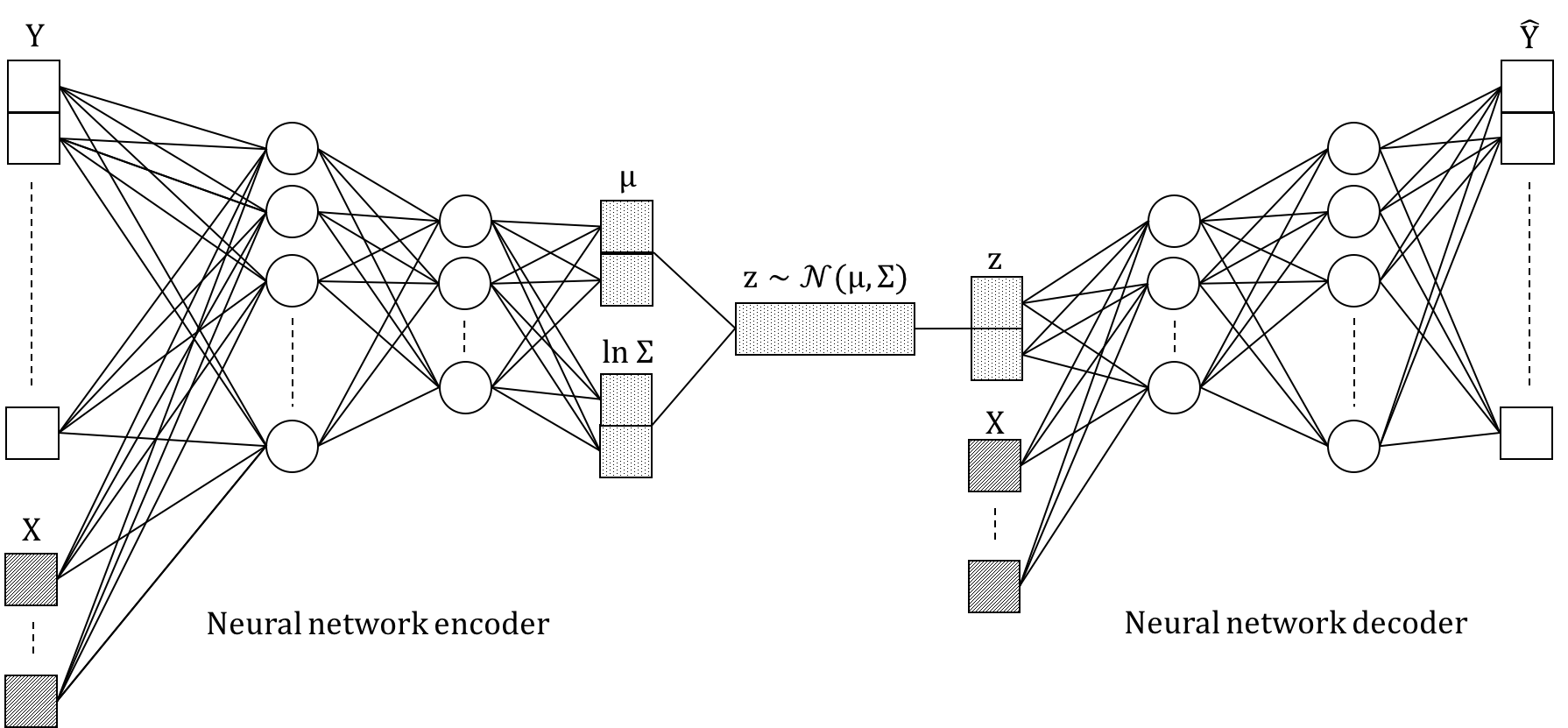} 
\caption{Diagram of a conditional variational autoencoder.}
\label{fig:cvae}
\end{figure}

Finally, conditional variational autoencoders (CVAE)~\cite{kingma2014semi} are an extension of VAE where a vector of exogenous variables $X$ is given as input to both the decoder and the encoder.
Adding this conditional information may improve the reconstruction. 
Figure~\ref{fig:cvae} depicts a scheme of the CVAE architecture used in the experiments.
The encoder takes as input a daily energy consumption profile $Y$ (so namely a $H$-vector gathering the half-hourly records of energy consumption) and an exogenous vectors $X$ (with calendar, weather, and tariffs information) and outputs the $d$-dimension vectors $\mu$ and $\log \Sigma$ (it is usual to consider a log-transformation, see~\cite{marot2019interpreting}).
The vector $\log \Sigma$ is also of dimension $d$. Indeed, only the diagonal of the covariance matrix $\Sigma$ is encoded since both approaches (diagonal and full-matrix) were tested and there was no major difference
on the reconstruction loss (obviously the regularization term is higher for a full covariance matrix). 
Since considering a full-matrix (which is symmetric) increases the dimension of encoders outputs (from $2d$ to $d(d+1)/2$) and the CVAE converges slower, we decided to keep a diagonal matrix to encode the covariance matrix $\Sigma$.
The random variable $Z$ is then sampled and given to the decoder as well as the vector of exogenous variables $X$. Finally, the decoder outputs $\widehat{Y}$.\\

Once the CVAE is trained, the decoder is isolated and used to generate data. For any day $s$, it is enough to sample a random variables $Z_s \sim \mathcal{N}(0,I_{d})$ in the latent space and give it as input to the decoder, combined with a vector of exogenous variables $X_s$ (that could be taken from the original data set or eventually created). Then, the decoder generates a $H$-vector $\widehat{Y}_s$ that corresponds to a new randomly generated daily consumption profile, for the day $s$ and the contextual variables $X_s$. 

\subsubsection{Implementation details}

The CVAE were implemented by using the software libraries~\texttt{Tensorflow} and \texttt{Keras} in \texttt{Python} programming language. 
The architecture of a CVAE is defined by the latent dimension $d$ as well as the number of layers and units in encoder and decoder neural networks. 
We use dense layers which are deeply connected neural network layers.
Once the architecture of the CVAEs is set, hyperparameters are chosen: the neural activation functions, the initialization method for neural weights and the parameter $\eta $, defined in Equation~\eqref{eq:loss}, that balances the two terms of the loss.
The choice of the architecture and hyper-parameters calibration is detailed in Section~\ref{sec:hyper_cal}.\\

In order to optimize a CVAE, so namely to compute weights and bias for each neural of both the encoder and the decoder, the loss is minimized by using the Adam optimizer (see~\cite{kingma2014adam}), an extension of stochastic gradient descent method, which is commonly used in deep learning and already implemented in \texttt{Keras}. Note that the learning rate of this optimizer is also an hyper-parameter to set before training CVAEs. \\

Finally, the energy consumption records are rescaled to get values between $0$ and $1$ by computing the maximum $Y_{\mathrm{max}}$ and minimum $Y_{\mathrm{min}}$ of the energy consumption observed on the train period. 
The generated value are re-scaled to get coherent profile, mostly between $Y_{\mathrm{min}}$ and $Y_{\mathrm{max}}$.\\

We recall that the data described in Section~\ref{sec:data} was divided into two data sets: the training set contains 75\% for the observations (sampled randomly from the complete data set) and is used to train the CVAE (see Table~\ref{tab:data_set}); the testing set, made with the remaining daily observations, is used to calibrate hyper-parameters (see Section~\ref{sec:hyper_cal}). Finally, as CVAE may converge into local minima, many CVAE are trained and the testing set is also used to select the best one (see Section~\ref{subsec:selec}).

\subsection{Hyper-parameters calibration}\label{sec:hyper_cal}

The process described below will be applied for each of the cluster defined in Section~\ref{sec:clustering}, for which a half-hourly energy consumption profile for each day of 2013 is available.

\subsubsection{Methodology} 

To perform CVAEs hyperparameter calibration we opt for a grid search approach that is simply an exhaustive searching through a manually specified subset of the hyperparameter space.
This optimization is guided by the performance metric detailed below, which is simply an evaluation on a held-out validation set. 
For each set of parameters, namely for each point of the grid, we train a CVAE and test it according to the procedure described below.
Once the CVAEs have converged, (we stop the convergence process when the loss is not decreasing any more), we compute the mean squared error (which corresponds to the reconstruction loss) on the testing set made of the observations $Y_{T_0+1},\dots,Y_{T}$:
\begin{gather}
\mathrm{MSE}  =  \frac{1}{T-T_0} \sum_{t=T_0+1}^{T} \big\| Y_t-\widehat{Y}_t \big\|^2\, \quad \mathrm{where} \quad \widehat{Y}_t= D(Z_t) \quad
 \mathrm{with} \quad Z_t\sim\mathcal{N}\big(\mu(Y_t),\Sigma(Y_t)\big)\,.
\end{gather}
The architecture and hyperparameters of the CVAE hat reaches to lowest MSE are kept.

\subsubsection{Results}

We tested different values from $1$ to $20$ for the latent dimension $d$ and reached a final value of $4$, which is coherent with the results in~\cite{marot2019interpreting} for the daily energy consumption in France.
Moreover, we also performed a principal component analysis (PCA) on the consumption data and found that $4$ components were enough to explain more 80\% of the variance in the data.
We tested CVAEs with one or two hidden layers of 10, 15, 20 or 25 units per layer and concluded that an architecture with a hidden layer of at least 15 neurons performed much better than smaller architectures.
We continued to increase the number of layers or the number of neurons per layer, but without improvement in the MSE. Moreover, the number of iterations necessary before convergence increased. So we decided to keep a single hidden layer of 15 units for both the encoder and the decoder. \\

Concerning the activation function of the neurons; rectified linear unit (ReLU), linear, and sigmoid functions were tested and there was no doubt that the best performance was obtained with a ReLU activation function. \\

For the initialization of the network weights, we compared various \texttt{Keras} initializers (Glorot uniform, HE normal, Lecun normal, Zeros, Ones) and a manual initialization with PCA (as described in~\cite{miranda2014optimizing}). We noticed that the weigths initialization does not have a strong impact on the results and therefore the Glorot uniform initializer was selected~\cite{glorot2010understanding}.\\

For the regularization parameter $\eta$ that balances the two terms (reconstruction and regularization) in the loss function, various strategies to tune its value already exist. For example, \cite{higgins2017beta} showed that a constant $\eta>1$ may outperform classical VAE (defined with $\eta=1$). Moreover,~\cite{liang2018variational} and~\cite{bowman2015generating} considered a moving parameter that gradually increases from $0$ to $1$ across iterations, linearly and according to a sigmoid, respectively. 
We tried the three approaches and opted for a constant regularization parameter equal to $10$.
Finally, we tested various learning rates for Adam optimizer but did not notice major variations in the performance, so we set it to $10^{-3}$.

\subsection{Conditional variables preprocessing}
\label{subsec:var} 
We tried various combinations of the exogenous variables described in Table~\ref{tab:var} and selected the one with the lowest MSE on the testing set. For a day $t$, the conditional vector $X_t$ gathers the variables described below. \\

Without loss of generality, prices are categorical variables (Low, Normal or High), so, for an day $t$ and an half-hour $h$, the prices $p_t^h$ are encoded into two binary variables $\mathbf{1}_{p_t^h=\mathrm{Low}},$ and $\mathbf{1}_{p_t^h=\mathrm{High}}$ (if these two variables are null in the same time, the tariff is Normal). The position-in-the-year $\kappa_t \in [0,1]$ and the binary variable $w_t$ for the type-of-day are also considered.\\

Taking into account the half-hourly temperature $\tau_t^1,\dots, \tau_t^{H}$ significantly improves the MSE on the testing set, but the dimension of the conditional variables vector is then quite high. We tried to reduce the dimension of the temperature profile and obtained better results. A PCA was performed on the vectors made of all temperatures at day $t$ (half-hourly records and smoothed temperature).
Three components were enough to explain 98\% of the variance.
Therefore, we only keep the three components provided by the PCA and re-scale them between $0$ and $1$ to provide the variables $\widetilde{\tau}_t^1,\widetilde{\tau}_t^2,\widetilde{\tau}_t^3$. Then, they are considered as conditional variables (the daily temperature profiles $(\tau_t^1,\dots, \tau_t^{H})$ are not taking into account anymore).\\

Therefore, for a day $t$, the vector of conditional variables $X_t$ is made of the continuous variables  $\widetilde{\tau}_t^1,\widetilde{\tau}_t^2,\widetilde{\tau}_t^3$, and $\kappa_t$ that lie in $[0,1]$and of the binary variables $w_t$, $\mathbf{1}_{p_t^1=\mathrm{Low}}, \dots, \mathbf{1}_{p_t^{H}=\mathrm{Low}}$, and $\mathbf{1}_{p_t^1=\mathrm{High}}, \dots, \mathbf{1}_{p_t^{H}=\mathrm{High}}$. 

\subsection{Simulator creation}
\label{subsec:selec}
Finally, we emphasize that CVAEs may converge into local minima.
To avoid it, each CVAE is trained $50$ times and the one with the lowest MSE on testing set is selected.
For each of the cluster presented in Section~\ref{sec:clustering}, we thus get a CVAE that takes as inputs the daily energy consumption profile $Y_t = (Y^1_t\dots,Y^{H}_t)$ of the considered cluster (which is rescaled during the training process) and the conditional vector $X_t$ described above. 
Then, the decoder is isolated and enables the generation of new data.
Indeed, for a new vector $X_{t'}$ at a day $t'$, which can either be created or extracted from the data test, we sample a vector $Z_{t'}\sim \mathcal{N}(0,I_{d})$ and give these two vectors as inputs of the decoder, which outputs a daily energy consumption profile. 
The quality of the generated data is evaluated in two situations.
First, samples for the conditional vectors $X_{T_0+1}, \dots, X_{T}$ associated with the training set are generated. Thus, we will measure the ability of the data generators to forecast energy consumption (we will see that we can deduce a foretasted density from the generated samples).
Secondly, we will create new vectors $X_{t}$ for which we modify the variables  $\mathbf{1}_{p_t^h=\mathrm{Low}},$ and $\mathbf{1}_{p_t^h=\mathrm{High}}$ in order to measure the impact of tariff changes. 
These results are presented in Section~\ref{sec:results} and compare them with data generated according to a semi-parametric data generator presented below.

\section{Semi-parametric generator: Additive Model}\label{sec:semi_param}

The following semi-parametric method based on generalized additive models (GAM), see~\cite{wood2006generalized}, is proposed to generate new daily consumption profile data.
GAMs form a powerful and efficient semi-parametric approach to model electricity consumption (see, among others, \cite{gaillard2016additive}) as a sum of independent exogenous variable effects.
Here, we assume that there exists a class of functions $\mathcal{F}$, such that, for a given half-hour $h$ and an instance $t$, with $x^h_t$ a vector of exogenous variables and $p^h_t$ the tariff, the energy consumption expectation satisfies
\begin{equation}
\E[Y_t^h]=f^h(x^h_t, p^h_t), \quad f^h \in \mathcal{F}.    
\end{equation}

After estimating the functions $f^h$ (we detail further the set $\mathcal{F}$ and how GAMs may approximate these functions), we could compute the residuals and try to fit a model on them. 
They are centered, but a time dependence is observed, so adding a independent white noise to each forecast will not provide realistic profiles. It is important to note that the same problem can be found in renewable energy uncertainty forecasting and the need to generate scenarios (or trajectories) with inter-temporal dependency structure for multi-period stochastic optimization (see~\cite{pinson2009probabilistic} for more details). \\

In this paper, we propose an approach based on a conjoint estimation of both mean and variation of the energy consumption. Then, we tried to used Gaussian copula to create trajectories, applying the methods proposed by Pinson \textit{et al.}~\cite{pinson2009probabilistic}. We faced an important problem: as soon as the function $f^h$ is not very well-estimated, the residuals variance comes, in majority, from the estimation error. 
More precisely, a bad estimation of the expected consumption leads to an increase of the estimated standard deviation.
\\

As the focus is on generating realistic a profile (and not necessary on having the best forecast in expectation), the standard deviation used to simulate data must reflect the variability observed in energy consumption data. 
Thanks to the causality model of Section~\ref{sec:causality}, that is now fitted on cluster consumptions (and not on individual ones), we can estimate the standard deviation of the noise as a function of the tariff and the half-hour $h$. We recall that we denote by $\sigma^h(p)$ the approximation of the standard $ \sqrt{ \Var [Y^h(i) \, | \, P=p ]}$ deviation associated with the half-hour $h$ and the tariff $p$  -- see Equation~\eqref{eq:var}.
It is used to normalize the residuals, which should then be centered and of variance $1$ (but not independent).
Finally, we consider the standardized residual vectors and compute an estimation of their correlation matrix $\Sigma$. 
We can now generate new data points this way:
\begin{gather}
\begin{bmatrix}
 Y_t^1\\
\vdots \\
Y_t^H 
\end{bmatrix}
=   \begin{bmatrix}
f^1 \big(x_t^1, p^1\big)  \\
\vdots \\
f^H \big(x_t^H,p^H \big) 
\end{bmatrix} 
+ \big(\sigma^1(p^1),\dots, \sigma^H(p^H) \big)^{\transp} E_t \,, 
\quad \mathrm{where} \quad  E_t \sim \mathcal{N}(0, \Sigma)\,.
\label{eq:semi_param_mod}
\end{gather}
Functions $(f^h)_{1\leq h \leq H}$ are estimated with GAMs and the exogenous vector $x_t^h$ gathers the temperature of the instance at the considerate half-hour $\tau_t^h$, the smoothed temperature $\bar{\tau}_t$, the position in the year $\kappa_t$, the binary variable $w_t$, which is equal to 1 if the day considered is a working day and 0 otherwise.
For each half-hour $h$, we set the same underlying GAM:
\begin{align}
f^h(x_t^h,p_t^h)=s^h_{\tau}(\tau^h_t) + s^h_{\bar{\tau}}(\bar{\tau_t}) + s^h_{\kappa}(\kappa_t) 
+\alpha^h w_t + \xi_{\mathrm{Low}}^h \mathbf{1}_{p_t^h=\mathrm{low}}+ \xi_{\mathrm{high}}^h\mathbf{1}_{p_t^h=\mathrm{High}} \,.
\label{eq:gam}
\end{align}
Therefore, $\mathcal{F}$ is the set of functions that can be written this way.
The  $s^h_{\tau}$, $s^h_{\bar{\tau}}$, and $s^h_{\kappa}$ functions  are catching the effect of the temperatures and of the yearly seasonality. They are approximated by cubic splines, i.e. C$^2$-smooth functions made up of sections of cubic polynomials joined together at points of a grid (the knots). Fixing the number of knots $k$ and their position is sufficient to determine a linear basis of dimension $k$ in which these functions can be projected. The \texttt{mgcv R-}package allows to estimate the coordinates of the splines in their basis and the coefficients $\alpha^h$, $\xi_{\mathrm{Low}}^h$, and $\xi_{\mathrm{High}}^h$ that catch day of the week and tariff effects. 
Appendix~\ref{app:semi_param} provides details on the estimation of the correlation matrix $\Sigma$, which makes it possible to model the correlations between the consumption profiles of two half-hours of the same day, whereas keeping a variance of the residuals that varies according to the half-hour and the price.

\section{Evaluation of the Data Generators} \label{sec:results}

\subsection{Evaluation Metrics}

By generating lots of energy consumption profiles from the simulators, an estimation of their densities can be obtained. Therefore, we use some proper scoring scores from probabilistic forecast evaluation to assess the quality of our generators. The three scores detailed below allow to evaluate the data generated on the testing period and compare both generators. 
For a day $t$ of the testing set, from the vector of exogenous variables $X_t$, both generators output $H$-random vectors that are assumed to be drawn from an underlying distribution $\widehat{F}_t$. These distributions approximate the true and unknown $H$-dimensional distributions $F_t$ from which the observation $(Y_t^1,\dots, Y_t^{H})$ is actually drawn. 
We generate $N=200$ samples $\widehat{Y}_t^{(1)}, \dots, \widehat{Y}_t^{(N)}$ for each generator.
From these $H$-random vectors, we can approximate the three scores described below, that measure the adequacy between the observation vectors $Y_t$ and the distributions $\widehat{F}_t$.\\

First of all, for a distribution $F$, and a vector of observation $y$, the root mean squared error is considered: 
$\mathrm{RMSE\,}(F,y) =  \big\| \mathbb{E}[Y] - y  \big\|$,
where $Y$ is a random vectors  distributed according to $F$.
The first score is thus the RMSE
 between the expectation of the distribution $\widehat{F}_t$ (which we approximate with empirical mean of the generated samples) and the observation $Y_t$:
\begin{equation}
\mathrm{RMSE\,}(\widehat{F}_t,Y_t)\approx  \Big\|  \frac{1}{N}\sum_{i=1}^N \widehat{Y}_t^{(i)} - Y_t \Big\|\,.
\end{equation}
Here, the expectation of the distribution $\widehat{F}_t$ is actually seen as a forecast of the energy consumption $Y_t$. 
But to evaluate the quality of $\widehat{F}_t$,  a criterion including the variance and shape of the densities is necessary. \\

The two other scores are proper scoring rules used to evaluate weather ensembles or temporal trajectories generated by a statistical method (e.g., copula model). The energy score, introduced by Gneiting and Raftery~\cite{gneiting2007strictly}, generalizes the univariate continuous ranked probability score (CRPS) and is defined as
\begin{equation}
\mathrm{EN\,}(F,y)=\mathbb{E} \Big[ \big\| Y- y \big\| \Big] - \frac{1}{2}\, \mathbb{E} \Big[ \big\| Y -  Y' \big\| \Big] \,,
\end{equation}
where $Y$ and $Y'$ are two independent random vectors that are distributed according to $F$.
This score is approximated by splitting the generated samples in two groups  
$\widehat{Y}_t^{(1)}, \dots, \widehat{Y}_t^{(N/2)}$ and $\widehat{Y}_t^{(N/2+1)}, \dots, \widehat{Y}_t^{(N)}$:
\begin{align}
\mathrm{EN\,}(\widehat{F}_t,Y_t)\approx \frac{2}{N}\sum_{i=1}^{N/2} \Big\| \widehat{Y}_t^{(i)}- Y_t \Big\| \quad - \frac{1}{N} \sum_{n=1}^{N/2}  \Big\| \widehat{Y}_t^{(i)} - \widehat{Y}_t^{(N/2+i)} \Big\|\,.
\label{eq:energyscore_approx}
\end{align}

Scheuerer and Hamill have shown 
that the ability of energy score to detect correctly correlations between the components of the multivariate distribution was limited (see~\cite{scheuerer2015variogram} for further details). 
To remedy, they introduced the variogram score of order $p$:
\begin{align}
\mathrm{VG}_p(F,y) = &  \sum_{h,h'=1}^{H}  \bigg( \big|y^h - y^{h'}\big|^p \notag \\
&\quad - \mathbb{E} \Big[ \big|Y^h -Y^{h'} \big|^p \Big] \bigg)^2\,,
\label{eq:vg}
\end{align}
where $Y$ is a random vectors  distributed according to $F$.
On simulated data, they compared the performance of different scores (including the energy score) with the variogram scores for various $p$.
 This score is approximated with:
\begin{align}
\mathrm{VG}_p(\widehat{F}_t,Y_t)\approx 
 \sum_{h,h'=1}^{H}  \Bigg( \big|Y_t^h - Y_t^{h'}\big|^p -
\frac{1}{N}\sum_{i=1}^N 
  \bigg| \Big(\widehat{Y}_t^{(i)}\Big)^{h} - \Big(\widehat{Y}_t^{(i)}\Big)^{h'} \, \bigg|^p \Bigg)^2\,.
  \label{eq:vg_approx}
\end{align}

We emphasize that for all the scores above, the smaller the value, the better the forecast.

\subsection{Numerical results}

\begin{figure*}[t]
\centering
\includegraphics[width=0.32\textwidth]{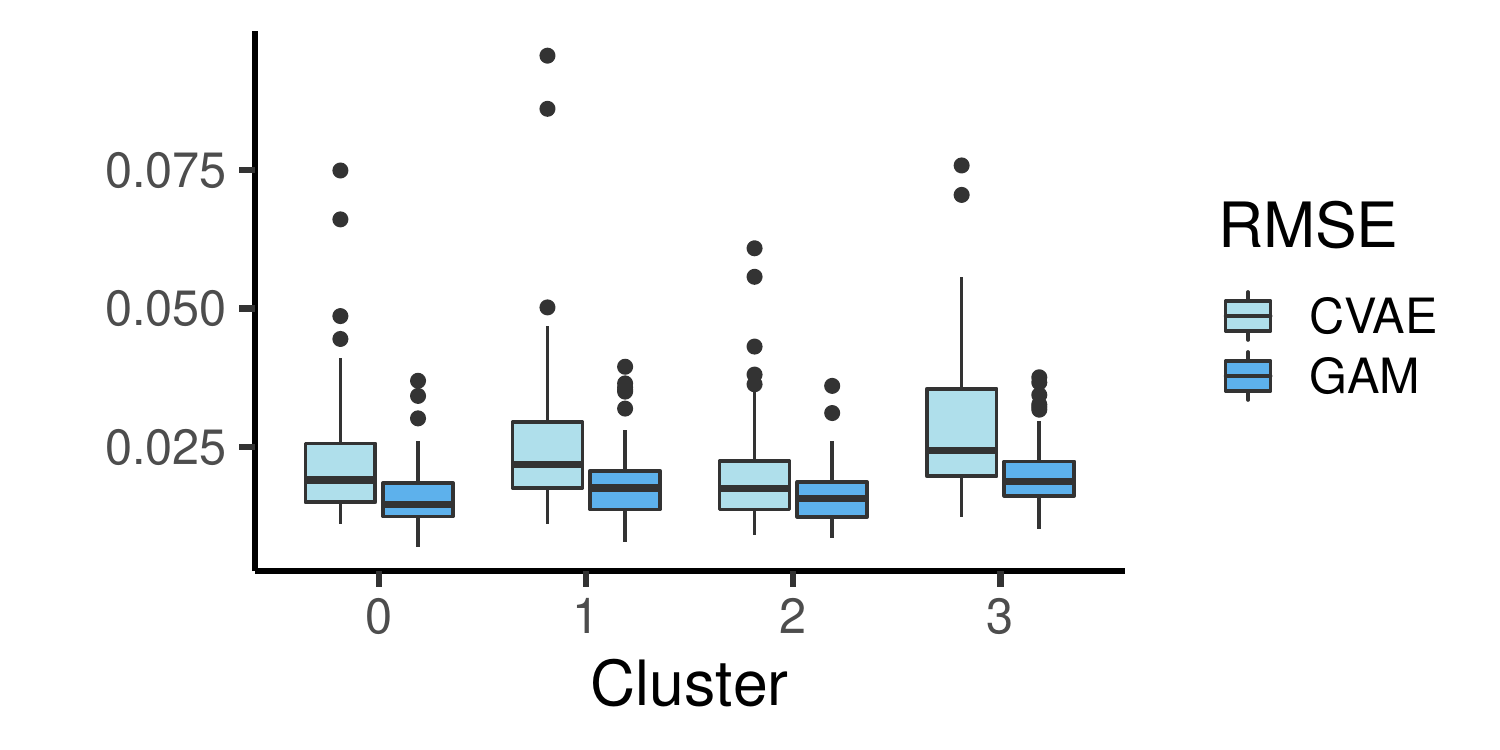}  \hfill
\includegraphics[width=0.32\textwidth]{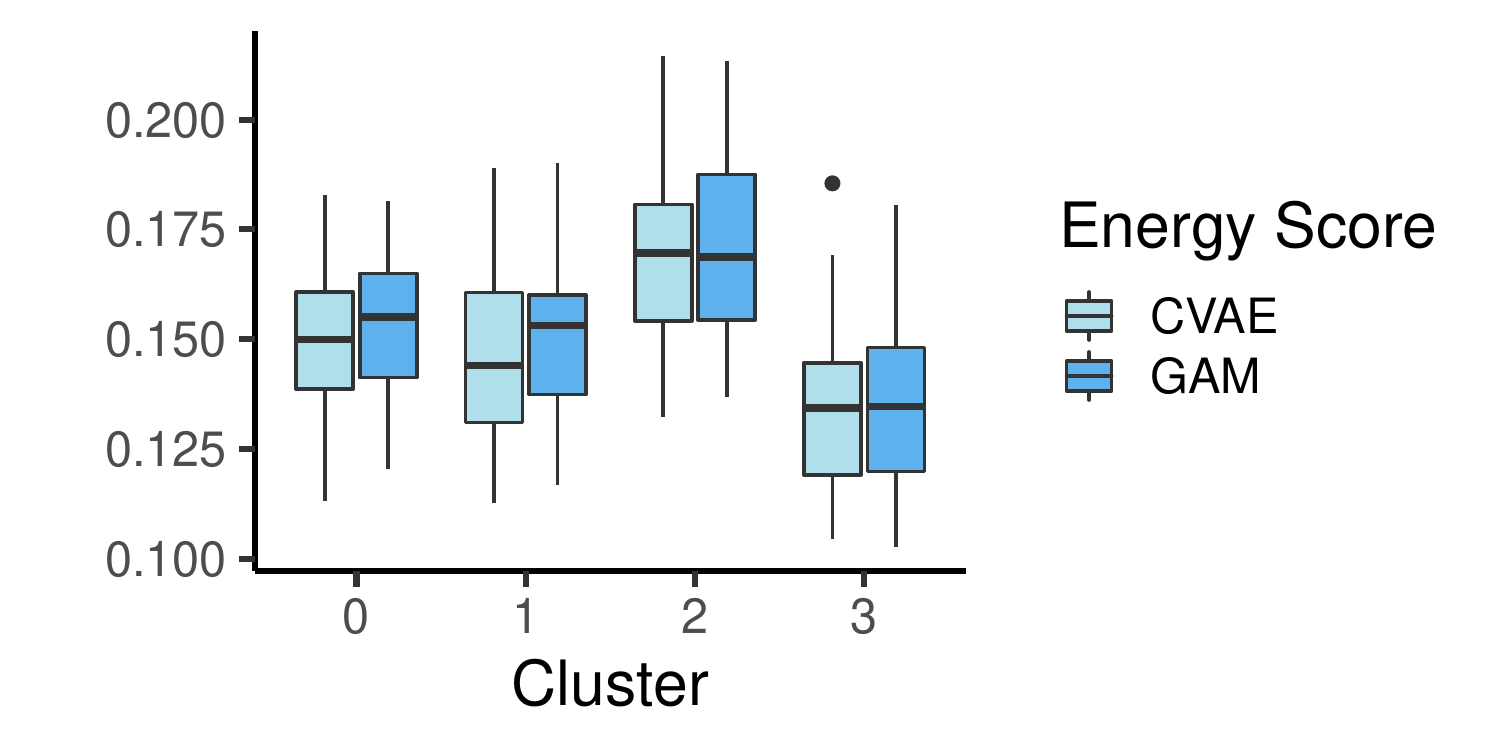} 
\includegraphics[width=0.32\textwidth]{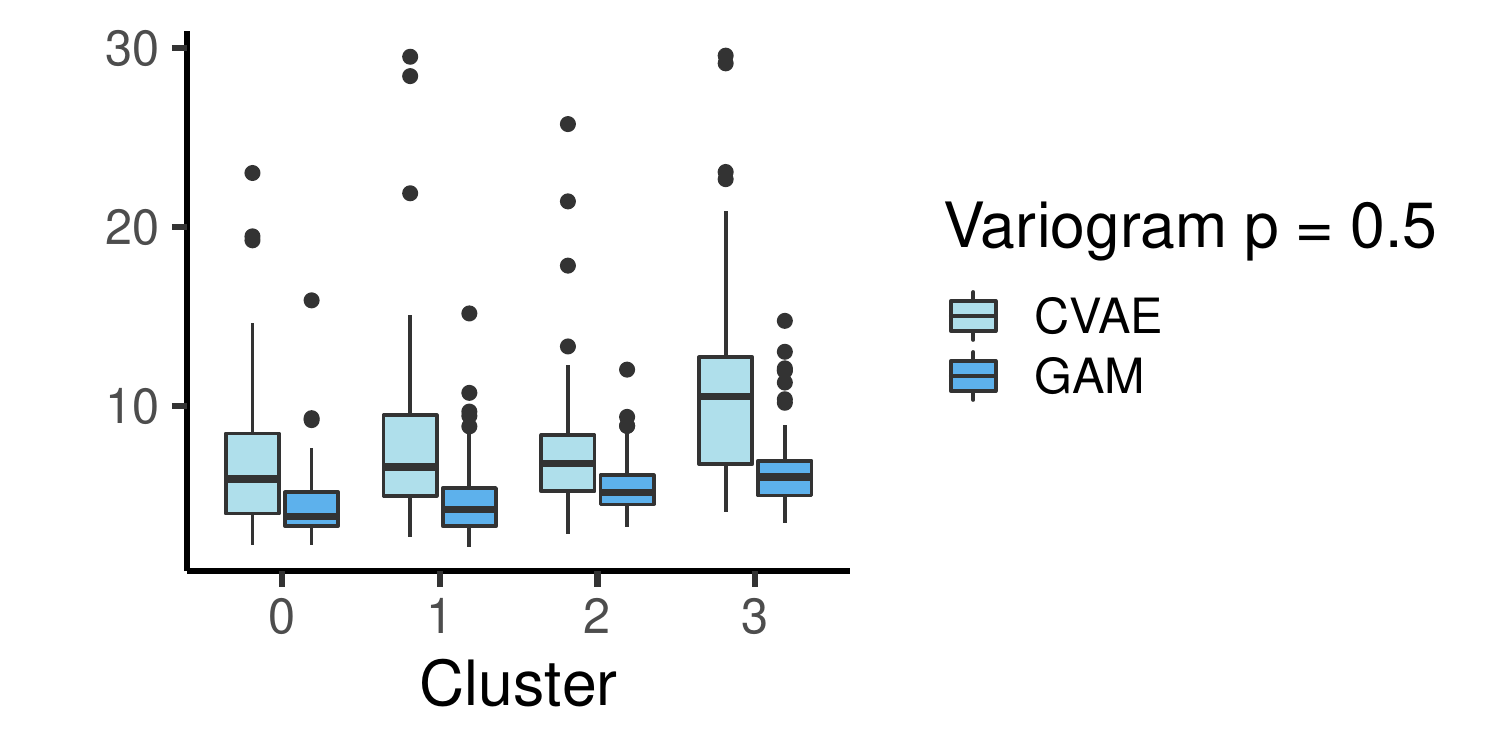} 
\caption{Boxplots. From left to right Root Mean Squared Error (RMSE), Energy Score and variogram for $p=0.5$ evaluated for each day of the data test set.}
\label{fig:scores}
\end{figure*}

For each cluster and each day $t$ of the testing set, we compute, for both generators (CVAE-based and GAM-based) the three scores (thanks to the $200$ generated samples). Results are represented by boxplots in Figure~\ref{fig:scores}.
Moreover, for the first three days of the testing set (that are actually the first three days of 2013), $20$ samples generated by the simulators for one of the $4$ clusters, their empirical means (computed on all the samples) and the corresponding observations $Y_t$ are plotted in Figure~\ref{fig:estimations}.
Plots of each cluster can be found in Figure~\ref{fig:estimations_all} of Appendix~\ref{app:graphs}.
\begin{figure*}[t]
\centering
\includegraphics[width=0.47\textwidth]{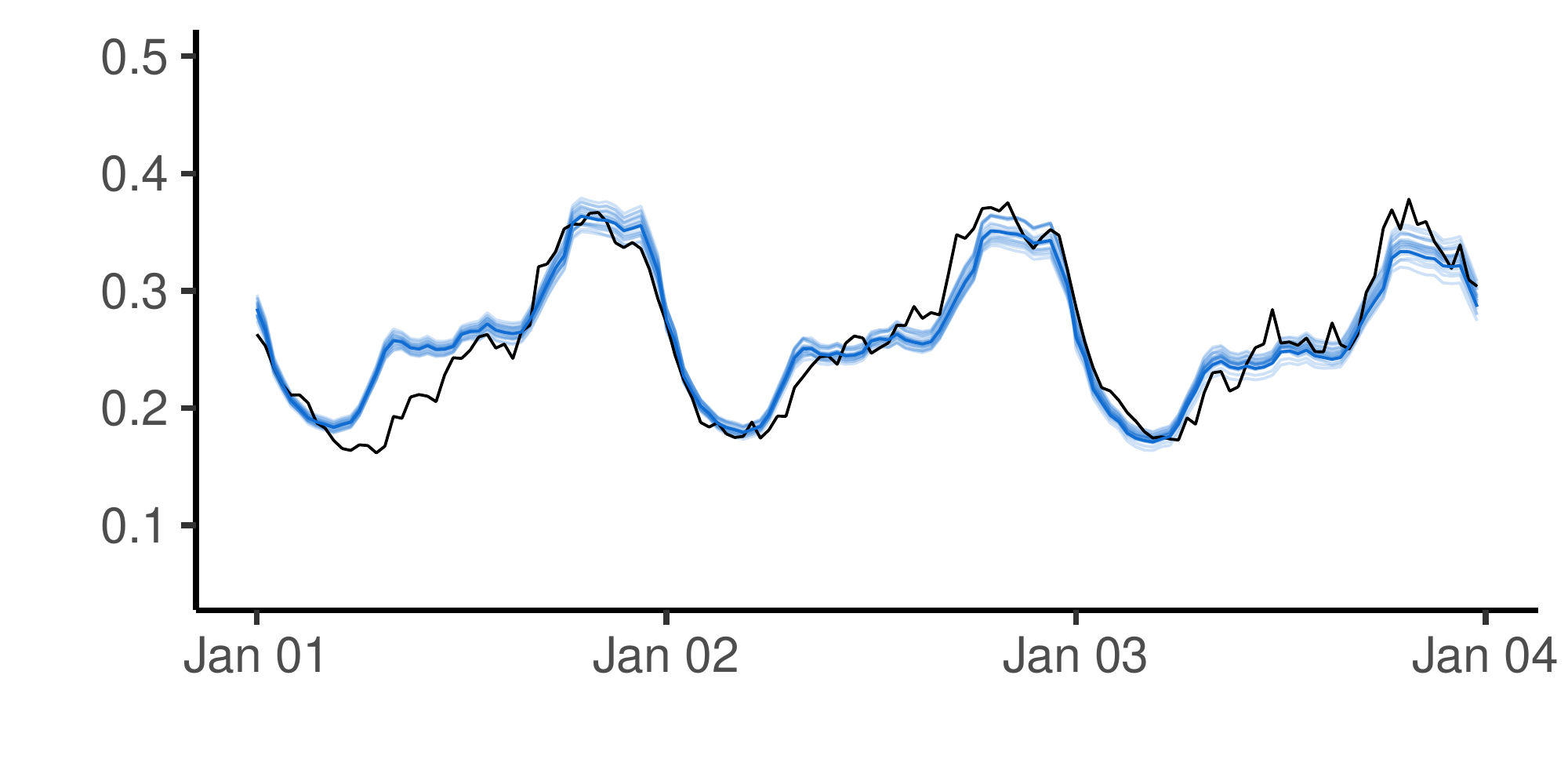}  \hfill
\includegraphics[width=0.52\textwidth]{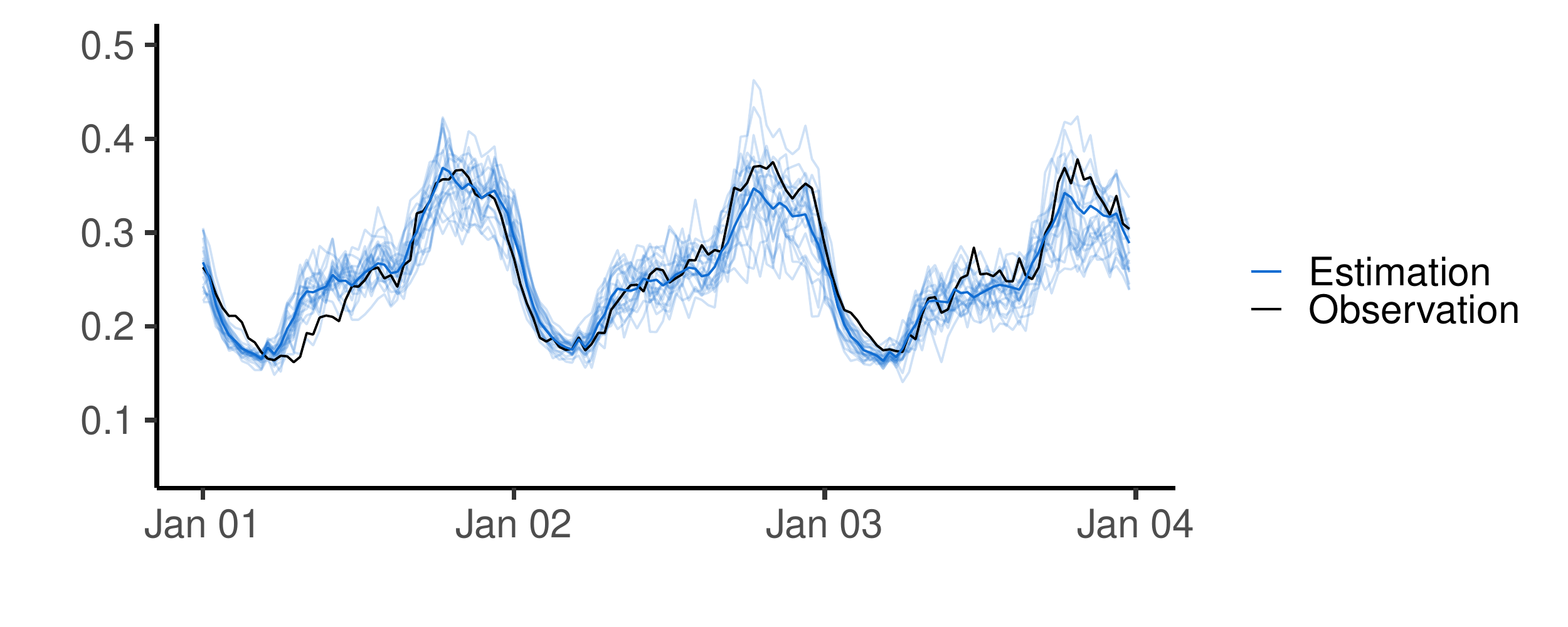} 
\caption{
Left: data generated with the CVAE-based generator.  Right: data generated with the GAM-based generator. 
Blue lines: for a single cluster over the first three days of the testing set, $20$ energy consumption profiles and empirical mean profile, calculated on $200$ samples (in bold), obtained by giving, to the two generators, the exogenous variables observed over this period.
Black line: real observed profiles.}
\label{fig:estimations}
\end{figure*}

It is quite difficult to discriminate significantly both generators from these scores, but some conclusions may still be drawn.
First, RMSE bloxplots and plots suggest that GAM-based generators work better than those that use CAVE when it comes to generating the average value of the original data (which is approximated by the empirical mean of the samples).
However, the energy score is slightly lower for the non-parametric approach (namely for CVAE-based simulator) than for the semi-parametric one (GAM-based simulator).
Thus, the method that consists in adding a noise term to a forecast in expectation may have some limits whereas CVAEs seem to catch correctly the distributions of daily energy consumption.\\

Experiments of~\cite{scheuerer2015variogram} highlight that, when the estimation of the average value of the original data is incorrect (namely when the expectation of $F$ differs from the expectation of $y$ in Equation~\eqref{eq:vg}), variogram scores increase. 
Moreover, a too low or a too high variance -- when the variance of $F$ differs from the one of $y$ -- also increases variogram.
Given the variogram scores and the plots, we conclude that CVAE-based generators face an estimation of expected energy consumption worst than the semi-parametric generator but provide also samples with a too low variance. 
Conversely, GAM-based generators provide sample with too much variance, which also leads to a quite high variagram score. \\

Moreover, in the CVAE approach, consumption values from an half-hour to another are very correlated, when in the semi-parametric one, consumption profiles are more erratic. 
Observations suggest that the real variances and correlations lie somewhere in between.
The semi-parametric method is very sensitive to the standard deviation $\sigma^h(p)$ estimations. Thus, over-estimating these variances, provide, for sure, very different samples, which may be also very erratic.
Concerning CVAE-based generator, the variance of the samples could manually be increased by generating the decoder inputs according to $\mathcal{N}(0,\sigma^2I_{d})$ with $\sigma>1$.\\

Finally, we emphasize that in the semi-parametric approach, the variance depends only on the tariff and on the half-hour, whereas in the CVAE, all exogenous variables are taking into account. Moreover, the next section presents some strong advantages of the CVAE generator.

\subsection{Impact of the tariff}

\begin{figure*}[t]
\centering
\includegraphics[width=0.47\textwidth]{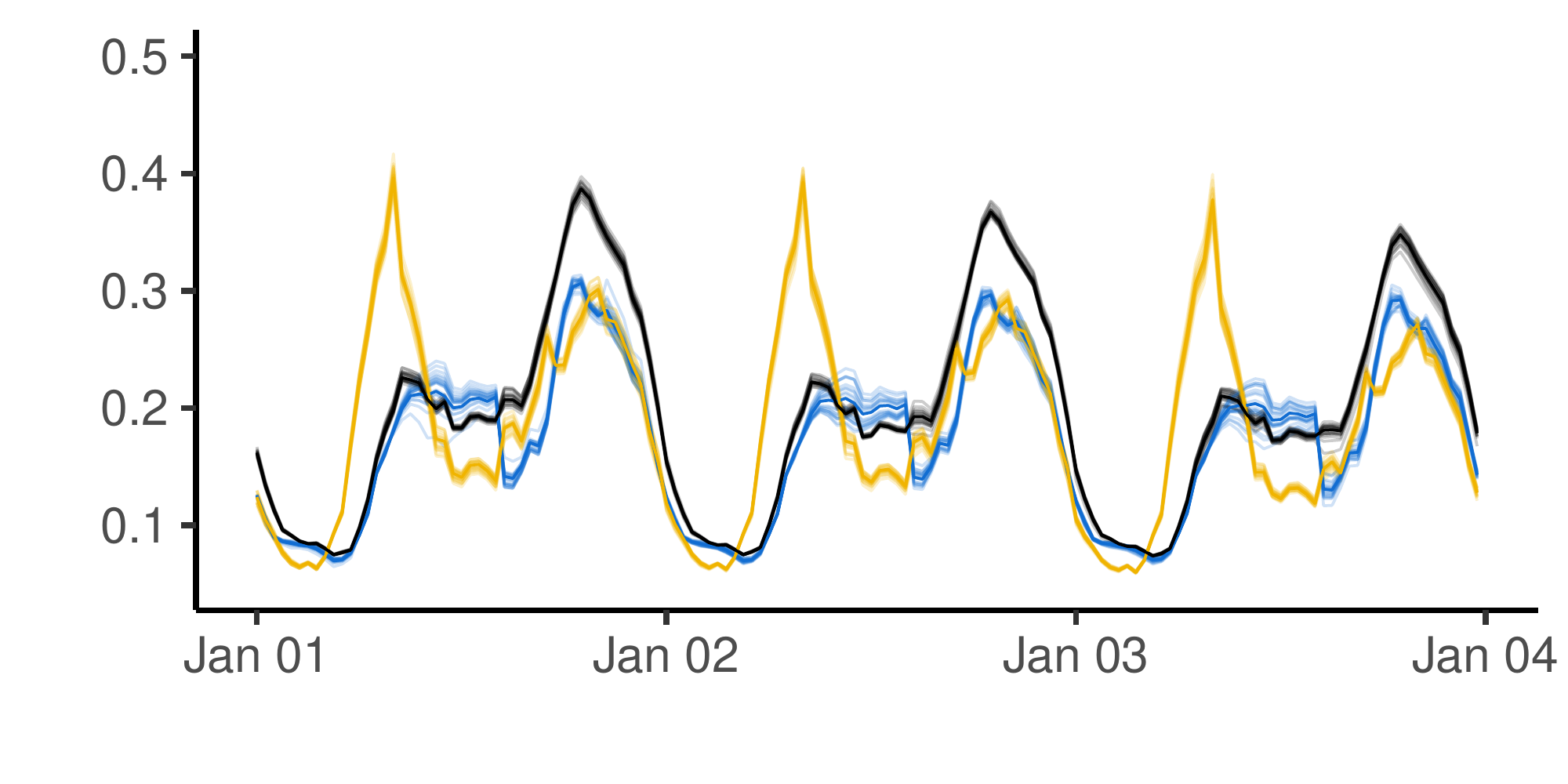}  \hfill
\includegraphics[width=0.52\textwidth]{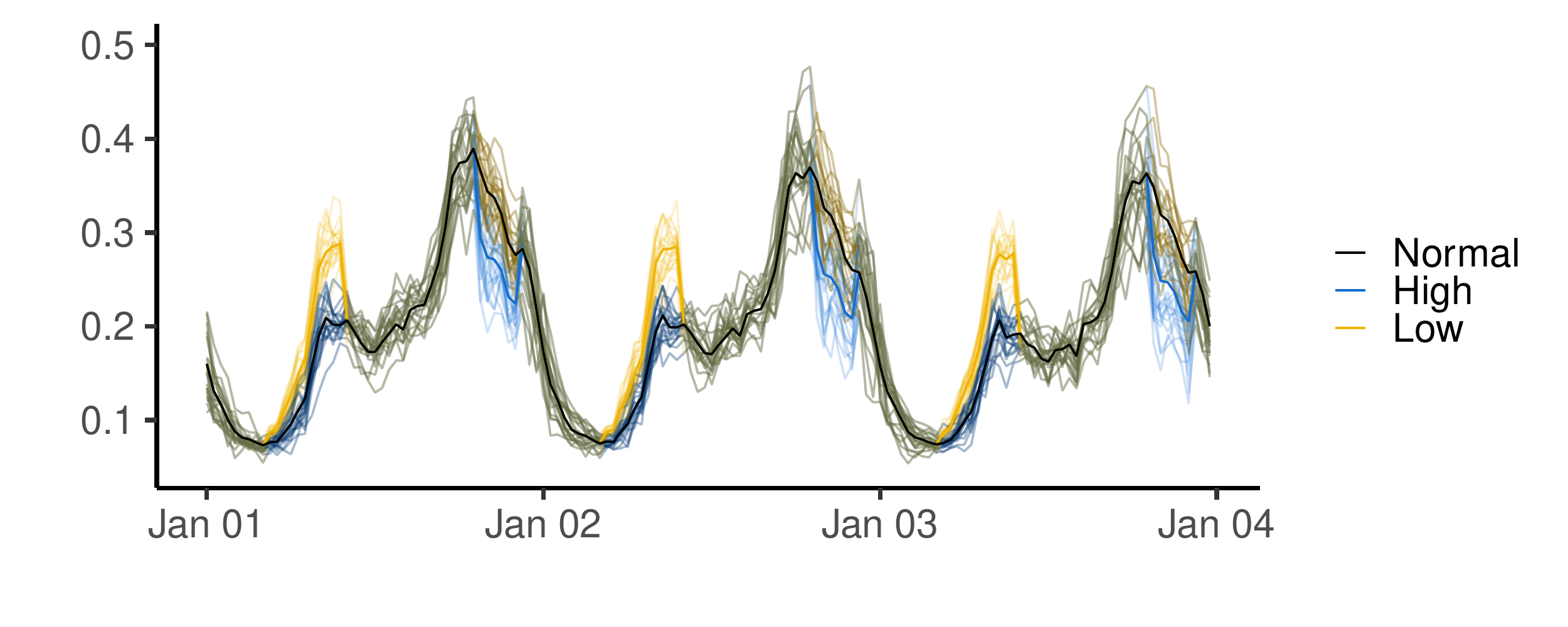} 
\caption{Left: data generated with the CVAE-based simulator.  Right: data generated with the GAM-based simulator.
Black lines: for a single cluster on the first three days of the data test set, $20$ energy consumption profiles and empirical mean profile, computed over $200$ samples (in bold), obtained by giving, to the two simulators, a Normal tariff for every half-hour and the weather and calendar variables observed over this period.
Blue lines: same plots but with a High tariff in the evening and Normal tariff otherwise.
Yellow lines: same plots but with a Low tariff in the early morning and Normal tariff otherwise.}
\label{fig:simulations}
\end{figure*}
In these last experiments, for a day $t$ of the testing set, three different conditional vectors $X_t^{_\mathrm{Normal}}$, $X_t^{_\mathrm{Low}}$ and $X_t^{_\mathrm{High}}$ are considered.
The tariff is Normal for all the day long for $X_t^{_\mathrm{Normal}}$. 
For the vector $X_t^{_\mathrm{Low}}$, Low tariff applies from 4:30 to 9:30 a.m., and Normal one otherwise, finally, tariff is Normal expect from 7:30 to 10 p.m. where it is High for  $X_t^{_\mathrm{High}}$.
For all other components, namely for the calendar and weather variables, $X_t^{_\mathrm{Normal}}$, $X_t^{_\mathrm{Low}}$, and $X_t^{_\mathrm{High}}$ are equal to $X_t$.
Still for the first three days of the testing set, $20$ samples generated by the generators for one of the $4$ clusters and their empirical means (computed with all the sample) are plotted in Figure~\ref{fig:simulations}.
Plots of each cluster can be found in Figure~\ref{fig:simulations_all} of Appendix~\ref{app:graphs}. \\

For both data generators, an increase of the consumption when tariff Low is applied and a decrease when the tariff is High are observed.
For the GAM-based generator, the effect of the tariff is very interpretable, it is actually measured by coefficients $\xi^h_{_\mathrm{Low}}$ and $\xi^h_{\mathrm{High}}$ of equation~\eqref{eq:gam}. This model makes actually this assumption that the tariff effect only depends on the half-hour.
Moreover, matrix $\Sigma$ models the correlations between the energy consumption at two half hours of the same day; this implicitly assumes that these correlations do not change according to the applied tariff profile.
Conversely, CVAE-based generator does not have this assumption and the effect of a tariff may differ from a day to another. \\

Moreover, two effects that cannot be modelled by the semi-parametric approach are observed.
First, the fall of the energy consumption occurs a little bit before the effective establishment of a special tariff High and continues a little after it is stopped. 
Thus, the effect of the High tariff exceeds the time window in which the special tariff is actually applied.
This is called a side effect.
Secondly, in comparison to a day of Normal tariff, when tariff Low is applied in the morning, there is a drop of the consumption in the afternoon and evening.
Similarly, we observe a little increase of the consumption in the afternoon when the tariff is High during the evening.  
Therefore, the fall or rise in consumption shifts to another time of the day when a special tariff is applied over a time window.
This is called a rebound effect.
These side and rebound effects are well known behaviors of consumers and it is very valuable that the generator detects them.
The main drawback of this non-parametric generator is the generation of non-intuitive consumption profiles when the input is a tariff profile never observed in the training set, like an entire day of High tariff for example.
This shows that the method has a limited generalization capacity.
Enlarging the data set, especially the variety of price signals, would eliminate this limitation.
On the other hand, for a full day of tariff High, the semi-parametric model generates samples with an energy consumption below the typical one for each half-hour, which is unrealistic since electricity uses cannot be delayed indefinitely.\\

Figure~\ref{fig:simulations_all} of Appendix~\ref{app:graphs} shows that tariff-responsiveness vary from a cluster to another, i.e., rebound or side effects are not always observed and the amount of electricity over or under consumed also depends on the considered cluster. These results fully illustrate the motivation behind the use of the causality model to cluster consumers.

\section{Conclusions}\label{sec:conclusions}

This paper proposed a data-driven and non-parametric methodology, based on CVAE, for generating synthetic energy consumption profiles for households enrolled in a DR program with different tariff schemes. The results for the largest data set publicly available (released by UK Power Networks) show that the proposed non-parametric generator captures correctly the effect of the exogenous variables and performs almost as well as the benchmark semi-parametric generator to generate the mean value of the original data. Besides and above all, the whole point of the CVAE approach comes from its ability to capture the effect of a daily tariff profile on the daily consumption profiles.
Indeed, unlike the semi-parametric generator that only captures the effect of a special tariff for the half-hours affected by this tariff change, the generator built from a decoder of a CVAE provides daily consumption samples for a daily tariff policy, including rebound and side effects. 
Moreover, for the same conditional variables as inputs, the generated samples differ from one group of consumers to another. Thus, the proposed clustering approach divides correctly the households according to their responsiveness to a tariff profile.\\

Finally, to deal with the lack of variability in the sent tariff profile of the original data set, we could imagine an online data generator: when a new tariff profile is sent, the observed consumption is integrated in the training set and the data generator is updated. The use of transfer learning methods could also improve the realism of the generated data. This machine learning field focuses on storing knowledge gained while solving one problem and applying it to a different but related problem.\\

Therefore, by combining data sets of consumer responsiveness to various DR programs (\textit{i.e.} by combining diverse knowledge of electricity demand in the face of tariff changes), a data set with a higher variability in the sent tariff profiles may be obtained. These data generators could be very useful to test potential future DR policies, before deploying such solutions in consumer households. Another topic of interest is the extension of the proposed model to consider privacy of the smart meter measurements and where recent research in privacy-preserving machine learning is a promising approach~\cite{AlRubaie2019}. 

\section*{Acknowledgments}

Margaux Br\'eg\`ere is very grateful to the entire team of the Center for Power and Energy Systems at INESC TEC for their wonderful welcome in Porto. In particular, she warmly thanks Kamalanathan Ganesan who explained to her the developed causality model and Jos\'e Ricardo Andrade for his help with Tensorflow and all the IT support.

\bibliographystyle{plainnat}
\addcontentsline{toc}{section}{Bibliographie}
\bibliography{arXiv}

\begin{thebibliography}{53}
\providecommand{\natexlab}[1]{#1}
\providecommand{\url}[1]{\texttt{#1}}
\expandafter\ifx\csname urlstyle\endcsname\relax
  \providecommand{\doi}[1]{doi: #1}\else
  \providecommand{\doi}{doi: \begingroup \urlstyle{rm}\Url}\fi

\bibitem[UKd()]{UKdata2020}
URL
  \url{data.london.gov.uk/dataset/smartmeter-energy-use-data-in-london-households}.

\bibitem[tem()]{temp_data}
URL \url{www.noaa.gov}.

\bibitem[Al-Rubaie and Chang(2019)]{AlRubaie2019}
Mohammad Al-Rubaie and J.~Morris Chang.
\newblock Privacy-preserving machine learning: {T}hreats and solutions.
\newblock \emph{IEEE Security \& Privacy}, 17\penalty0 (2):\penalty0 49--58,
  2019.

\bibitem[Alfaverh et~al.(2020)Alfaverh, Dena{\"\i}, and Sun]{Alfaverh2020}
Fayiz Alfaverh, M.~Dena{\"\i}, and Yichuang Sun.
\newblock Demand response strategy based on reinforcement learning and fuzzy
  reasoning for home energy management.
\newblock \emph{IEEE Access}, 8:\penalty0 39310--39321, 2020.

\bibitem[Bowman et~al.(2016)Bowman, Vilnis, Vinyals, Dai, Jozefowicz, and
  Bengio]{bowman2015generating}
Samuel~R Bowman, Luke Vilnis, Oriol Vinyals, Andrew~M Dai, Rafal Jozefowicz,
  and Samy Bengio.
\newblock Generating sentences from a continuous space.
\newblock In \emph{Proceedings of The 20th SIGNLL Conference on Computational
  Natural Language Learning}, 2016.

\bibitem[Br{\'e}g{\`e}re and Huard(2020)]{bregere2020online}
Margaux Br{\'e}g{\`e}re and Malo Huard.
\newblock Online hierarchical forecasting for power consumption data.
\newblock \emph{arXiv preprint arXiv:2003.00585}, 2020.

\bibitem[Br{\'e}g{\`e}re et~al.(2019)Br{\'e}g{\`e}re, Gaillard, Goude, and
  Stoltz]{bregere2019target}
Margaux Br{\'e}g{\`e}re, Pierre Gaillard, Yannig Goude, and Gilles Stoltz.
\newblock Target tracking for contextual bandits: Application to demand side
  management.
\newblock In \emph{Proceedings of the 36th International Conference on Machine
  Learning (ICML), PMLR 97}, pages 754--763, 2019.

\bibitem[Cali{\'n}ski and Harabasz(1974)]{calinski1974dendrite}
Tadeusz Cali{\'n}ski and Jerzy Harabasz.
\newblock A dendrite method for cluster analysis.
\newblock \emph{Communications in Statistics-theory and Methods}, 3\penalty0
  (1):\penalty0 1--27, 1974.

\bibitem[Capasso et~al.(1994)Capasso, Grattieri, Lamedica, and
  Prudenzi]{Capasso1994}
A.~Capasso, W.~Grattieri, R.~Lamedica, and A.~Prudenzi.
\newblock A bottom-up approach to residential load modeling.
\newblock \emph{IEEE Transactions on Power Systems}, 9\penalty0 (2):\penalty0
  957--964, 1994.

\bibitem[Chen et~al.(2018)Chen, Wang, and Zhang]{chen2018unsupervised}
Yize Chen, Xiyu Wang, and Baosen Zhang.
\newblock An unsupervised deep learning approach for scenario forecasts.
\newblock In \emph{2018 Power Systems Computation Conference (PSCC)}, pages
  1--7. IEEE, 2018.

\bibitem[Chicco et~al.(2006)Chicco, Napoli, and
  Piglione]{chicco2006comparisons}
Gianfranco Chicco, Roberto Napoli, and Federico Piglione.
\newblock Comparisons among clustering techniques for electricity customer
  classification.
\newblock \emph{IEEE Transactions on Power Systems}, 21\penalty0 (2):\penalty0
  933--940, 2006.

\bibitem[Dutta and Mitra(2017)]{Dutta2017}
G.~Dutta and Krishnendranath Mitra.
\newblock A literature review on dynamic pricing of electricity.
\newblock \emph{Journal of the Operational Research Society}, 68:\penalty0
  1131--1145, 2017.

\bibitem[Fidalgo et~al.(2012)Fidalgo, Matos, and Ribeiro]{Fidalgo2012}
José~Nuno Fidalgo, Manuel~António Matos, and Luis Ribeiro.
\newblock A new clustering algorithm for load profiling based on billing data.
\newblock \emph{Electric Power Systems Research}, 82\penalty0 (1):\penalty0
  27--33, 2012.

\bibitem[Gaillard et~al.(2016)Gaillard, Goude, and
  Nedellec]{gaillard2016additive}
Pierre Gaillard, Yannig Goude, and Rapha{\"e}l Nedellec.
\newblock Additive models and robust aggregation for {GEFCom2014} probabilistic
  electric load and electricity price forecasting.
\newblock \emph{International Journal of Forecasting}, 32\penalty0
  (3):\penalty0 1038--1050, 2016.

\bibitem[Ganesan et~al.(2019)Ganesan, Saraiva, and Bessa]{ganesan2019using}
Kamalanathan Ganesan, Jo{\~a}o~Tom{\'e} Saraiva, and Ricardo~J Bessa.
\newblock On the use of causality inference in designing tariffs to implement
  more effective behavioral demand response programs.
\newblock \emph{Energies}, 12\penalty0 (14):\penalty0 2666, 2019.

\bibitem[Glorot and Bengio(2010)]{glorot2010understanding}
Xavier Glorot and Yoshua Bengio.
\newblock Understanding the difficulty of training deep feedforward neural
  networks.
\newblock In \emph{13th International Conference on Artificial Intelligence and
  Statistics (AISTATS)}, pages 249--256, 2010.

\bibitem[Gneiting and Raftery(2007)]{gneiting2007strictly}
Tilmann Gneiting and Adrian~E Raftery.
\newblock Strictly proper scoring rules, prediction, and estimation.
\newblock \emph{Journal of the American statistical Association}, 102\penalty0
  (477):\penalty0 359--378, 2007.

\bibitem[Gottwalt et~al.(2011)Gottwalt, Ketter, Block, Collins, and
  Weinhardt]{Gottwalt2011}
Sebastian Gottwalt, Wolfgang Ketter, Carsten Block, John Collins, and Christof
  Weinhardt.
\newblock Demand side management--{A} simulation of household behavior
  undervariable prices.
\newblock \emph{Energy Policy}, 39:\penalty0 8163--8174, 2011.

\bibitem[Goude et~al.(2014)Goude, Nedellec, and Kong]{goude2014local}
Yannig Goude, Raphael Nedellec, and Nicolas Kong.
\newblock Local {{short}} and {{middle term electricity load forecasting}} with
  {{semi}}-{{parametric additive models}}.
\newblock \emph{IEEE Transactions on Smart Grid}, 5\penalty0 (1):\penalty0
  440--446, 2014.

\bibitem[Higgins et~al.(2017)Higgins, Matthey, Pal, Burgess, Glorot, Botvinick,
  Mohamed, and Lerchner]{higgins2017beta}
Irina Higgins, Loic Matthey, Arka Pal, Christopher Burgess, Xavier Glorot,
  Matthew Botvinick, Shakir Mohamed, and Alexander Lerchner.
\newblock beta-{VAE}: {L}earning basic visual concepts with a constrained
  variational framework.
\newblock \emph{5th International Conference on Learning Representations
  (ICLR)}, 2\penalty0 (5):\penalty0 6, 2017.

\bibitem[Hinton and Zemel(1994)]{hinton1994autoencoders}
Geoffrey~E Hinton and Richard~S Zemel.
\newblock Autoencoders, minimum description length and helmholtz free energy.
\newblock In \emph{NIPS'93: Proceedings of the 6th International Conference on
  Neural Information Processing Systems}, pages 3--10, 1994.

\bibitem[Iwafune et~al.(2020)Iwafune, Ogimoto, Kobayashi, and
  Murai]{Iwafune2020}
Yumiko Iwafune, Kazuhiko Ogimoto, Yuki Kobayashi, and Kensuke Murai.
\newblock Driving simulator for electric vehicles using the {M}arkov chain
  monte carlo method and evaluation of the demand response effect in
  residential houses.
\newblock \emph{IEEE Access}, 8:\penalty0 47654--47663, 2020.

\bibitem[Karlsen et~al.(2020)Karlsen, Hamdy, and Attia]{Karlsen2020}
Sophie~Schönfeldt Karlsen, Mohamed Hamdy, and Shady Attia.
\newblock Methodology to assess business models of dynamic pricing tariffs in
  all-electric houses.
\newblock \emph{Energy and Buildings}, 207:\penalty0 109586, 2020.

\bibitem[Kaufman and Rousseeuw(1987)]{kaufman1987clustering}
Leonard Kaufman and Peter~J Rousseeuw.
\newblock Clustering by means of medoids. statistical data analysis based on
  the l1 norm.
\newblock \emph{Y. Dodge, Ed}, pages 405--416, 1987.

\bibitem[Kingma and Welling(2014)]{kingma2013auto}
D.~P. Kingma and M.~Welling.
\newblock Auto-encoding variational bayes.
\newblock In \emph{2nd International Conference on Learning Representations
  (ICLR2014)}, 2014.

\bibitem[Kingma and Ba(2015)]{kingma2014adam}
Diederik~P Kingma and Jimmy Ba.
\newblock Adam: {A} method for stochastic optimization.
\newblock In \emph{3rd International Conference on Learning Representations
  (ICLR2015)}, 2015.

\bibitem[Kingma et~al.(2014)Kingma, Mohamed, Rezende, and
  Welling]{kingma2014semi}
Durk~P Kingma, Shakir Mohamed, Danilo~Jimenez Rezende, and Max Welling.
\newblock Semi-supervised learning with deep generative models.
\newblock In \emph{28th Conference on Neural Information Processing Systems
  (NIPS)}, pages 3581--3589, 2014.

\bibitem[Le~Ray and Pinson(2019)]{le2019online}
Guillaume Le~Ray and Pierre Pinson.
\newblock Online adaptive clustering algorithm for load profiling.
\newblock \emph{Sustainable Energy, Grids and Networks}, 17:\penalty0 100181,
  2019.

\bibitem[Le~Ray et~al.(2016)Le~Ray, Larsen, and Pinson]{le2016evaluating}
Guillaume Le~Ray, Emil~Mahler Larsen, and Pierre Pinson.
\newblock Evaluating price-based demand response in practice?with application
  to the ecogrid eu experiment.
\newblock \emph{IEEE Transactions on Smart Grid}, 9\penalty0 (3):\penalty0
  2304--2313, 2016.

\bibitem[Lee and Seung(1999)]{lee1999learning}
Daniel~D Lee and H~Sebastian Seung.
\newblock Learning the parts of objects by non-negative matrix factorization.
\newblock \emph{Nature}, 401\penalty0 (6755):\penalty0 788, 1999.

\bibitem[Li et~al.(2019)Li, Lei, Huang, Qin, and Chen]{Li2019}
Jinghua Li, Yongsheng Lei, Qian Huang, Zhijun Qin, and Bo~Chen.
\newblock Feature analysis of generalized load patterns considering active load
  response to real-time pricing.
\newblock \emph{IEEE Access}, 7:\penalty0 119443--119453, 2019.

\bibitem[Liang et~al.(2018)Liang, Krishnan, Hoffman, and
  Jebara]{liang2018variational}
Dawen Liang, Rahul~G Krishnan, Matthew~D Hoffman, and Tony Jebara.
\newblock Variational autoencoders for collaborative filtering.
\newblock In \emph{Proceedings of the 2018 World Wide Web Conference}, pages
  689--698, 2018.

\bibitem[L\'opez et~al.(2019)L\'opez, Pouresmaeil, Ca{\~n}izares, Bhattacharya,
  Mosaddegh, and Solanki]{Lopez2019}
Juan Miguel~Gonzalez L\'opez, Edris Pouresmaeil, Claudio~A. Ca{\~n}izares,
  Kankar Bhattacharya, Abolfazl Mosaddegh, and Bharatkumar~V. Solanki.
\newblock Smart residential load simulator for energy management in smart
  grids.
\newblock \emph{IEEE Transactions on Industrial Electronics}, 66\penalty0
  (2):\penalty0 1443--1452, 2019.

\bibitem[Mallet et~al.(2014)Mallet, Granstrom, Hallberg, Lorenz, and
  Mandatova]{mallet14}
P.~Mallet, P.~O. Granstrom, P.~Hallberg, G.~Lorenz, and P.~Mandatova.
\newblock Power to the people! european perspectives on the future of electric
  distribution.
\newblock \emph{IEEE Power and Energy Magazine}, 12\penalty0 (2):\penalty0
  51--64, 2014.

\bibitem[Marot et~al.(2019)Marot, Rosin, Crochepierre, Donnot, Pinson, and
  Boudjeloud-Assala]{marot2019interpreting}
Antoine Marot, Antoine Rosin, Laure Crochepierre, Benjamin Donnot, Pierre
  Pinson, and Lydia Boudjeloud-Assala.
\newblock Interpreting atypical conditions in systems with deep conditional
  autoencoders: the case of electrical consumption.
\newblock In \emph{ECML PKDD 2019 - European Conference on Machine Learning and
  Principles and Practice of Knowledge Discovery in Databases}, 2019.

\bibitem[Miranda et~al.(2014)Miranda, da~Hora~Martins, and
  Palma]{miranda2014optimizing}
Vladimiro Miranda, Joana da~Hora~Martins, and Vera Palma.
\newblock Optimizing large scale problems with metaheuristics in a reduced
  space mapped by autoencoders--{A}pplication to the wind-hydro coordination.
\newblock \emph{IEEE Transactions on Power Systems}, 29\penalty0 (6):\penalty0
  3078--3085, 2014.

\bibitem[Mohajeryami et~al.(2016)Mohajeryami, Moghaddam, Doostan, Vatani, and
  Schwarz]{Mohajeryami2016}
Saeed Mohajeryami, Iman~N. Moghaddam, Milad Doostan, Behdad Vatani, and Peter
  Schwarz.
\newblock A novel economic model for price-based demand response.
\newblock \emph{Electric Power Systems Research}, 135:\penalty0 1--9, 2016.

\bibitem[Muratori et~al.(2013)Muratori, Roberts, Sioshansi, Marano, and
  Rizzoni]{Muratori2013}
Matteo Muratori, Matthew~C. Roberts, Ramteen Sioshansi, Vincenzo Marano, and
  Giorgio Rizzoni.
\newblock A highly resolved modeling technique to simulate residential power
  demand.
\newblock \emph{Applied Energy}, 107:\penalty0 465--473, 2013.

\bibitem[O'Neill et~al.(2010)O'Neill, Levorato, Goldsmith, and
  Mitra]{oneill2010residential}
Daniel O'Neill, Marco Levorato, Andrea Goldsmith, and Urbashi Mitra.
\newblock Residential demand response using reinforcement learning.
\newblock In \emph{2010 First IEEE international conference on smart grid
  communications}, pages 409--414. IEEE, 2010.

\bibitem[Paatero and Tapper(1994)]{paatero1994positive}
Pentti Paatero and Unto Tapper.
\newblock Positive {{matrix factorization}}: {{a non}}-{{negative factor
  model}} with {{optimal utilization}} of {{error estimates}} of {{data
  values}}.
\newblock \emph{Environmetrics}, 5\penalty0 (2):\penalty0 111--126, 1994.

\bibitem[Park et~al.(2010)Park, Kim, Moon, Heo, and Yoon]{Park2010}
Seunghyun Park, Hanjoo Kim, Hichan Moon, Jun Heo, and Sungroh Yoon.
\newblock Concurrent simulation platform for energy-aware smart metering
  systems.
\newblock \emph{IEEE Transactions on Consumer Electronics}, 56\penalty0
  (3):\penalty0 1918--1926, 2010.

\bibitem[Pinson et~al.(2009)Pinson, Madsen, Nielsen, Papaefthymiou, and
  Kl{\"o}ckl]{pinson2009probabilistic}
Pierre Pinson, Henrik Madsen, Henrik~Aa Nielsen, George Papaefthymiou, and
  Bernd Kl{\"o}ckl.
\newblock From probabilistic forecasts to statistical scenarios of short-term
  wind power production.
\newblock \emph{Wind Energy}, 12\penalty0 (1):\penalty0 51--62, 2009.

\bibitem[Richardson et~al.(2010)Richardson, Thomson, Infield, and
  Clifford]{Richardson2010}
Ian Richardson, Murray Thomson, David Infield, and Conor Clifford.
\newblock Domestic electricity use: {A} high-resolution energy demand model.
\newblock \emph{Energy and Buildings}, 42\penalty0 (10):\penalty0 1878--1887,
  2010.

\bibitem[Rodrigues et~al.(2008)Rodrigues, Gama, and Pedroso]{Rodrigues2008}
Pedro~Pereira Rodrigues, João Gama, and Joao Pedroso.
\newblock Hierarchical clustering of time-series data streams.
\newblock \emph{IEEE Transactions on Knowledge and Data Engineering},
  20\penalty0 (5):\penalty0 615--627, 2008.

\bibitem[Rumelhart et~al.(1986)Rumelhart, Hinton, and
  Williams]{rumelhart1986learning}
David~E Rumelhart, Geoffrey~E Hinton, and Ronald~J Williams.
\newblock Learning representations by back-propagating errors.
\newblock \emph{Nature}, 323\penalty0 (6088):\penalty0 533--536, 1986.

\bibitem[Saez-Gallego and Morales(2017)]{saez2017short}
Javier Saez-Gallego and Juan~M Morales.
\newblock Short-term forecasting of price-responsive loads using inverse
  optimization.
\newblock \emph{IEEE Transactions on Smart Grid}, 9\penalty0 (5):\penalty0
  4805--4814, 2017.

\bibitem[Scheuerer and Hamill(2015)]{scheuerer2015variogram}
Michael Scheuerer and Thomas~M Hamill.
\newblock Variogram-based proper scoring rules for probabilistic forecasts of
  multivariate quantities.
\newblock \emph{Monthly Weather Review}, 143\penalty0 (4):\penalty0 1321--1334,
  2015.

\bibitem[Schofield et~al.(2014)Schofield, Carmichael, Tindemans, Woolf, Bilton,
  and Strbac]{schofield2014residential}
J~Schofield, R~Carmichael, S~Tindemans, Matt Woolf, M~Bilton, and G~Strbac.
\newblock Residential consumer responsiveness to time-varying pricing.
\newblock Technical report, Imperial College London, Report A3 for the ``Low
  Carbon London'' LCNF project, 2014.

\bibitem[Siano(2014)]{SIANO2014461}
Pierluigi Siano.
\newblock Demand response and smart grids -- {A} survey.
\newblock \emph{Renewable and Sustainable Energy Reviews}, 30:\penalty0
  461--478, 2014.

\bibitem[Sun et~al.(2017)Sun, Konstantelos, and Strbac]{Sun2017}
Mingyang Sun, Ioannis Konstantelos, and Goran Strbac.
\newblock {C}-vine copula mixture model for clustering of residential
  electrical load pattern data.
\newblock \emph{IEEE Transactions on Power Systems}, 32\penalty0 (3):\penalty0
  2382--2393, 2017.

\bibitem[Taylor(2003)]{Taylor2003}
J~W Taylor.
\newblock Short-term electricity demand forecasting using double seasonal
  exponential smoothing.
\newblock \emph{Journal of the Operational Research Society}, 54\penalty0
  (8):\penalty0 799--805, 2003.

\bibitem[Wood(2006)]{wood2006generalized}
Simon Wood.
\newblock \emph{Generalized Additive Models: An Introduction with {R}}.
\newblock CRC Press, 2006.

\bibitem[Yan et~al.(2013)Yan, Qian, Sharif, and Tipper]{Yan13}
Y.~Yan, Y.~Qian, H.~Sharif, and D.~Tipper.
\newblock A survey on smart grid communication infrastructures: Motivations,
  requirements and challenges.
\newblock \emph{IEEE Communications Surveys Tutorials}, 15\penalty0
  (1):\penalty0 5--20, 2013.

\end{thebibliography}
\newpage
\appendix 
 \section{}
 \subsection{Exponential smoothing}\label{app:exp_smooth}

This section describes the construction of the exogenous variable $\bar{\tau}_t$, which is a smoothed air temperature (that models the thermal inertia of buildings, with the $a$-exponential smoothing. To do so, London temperatures are considered as a $1$-dimensional half-hourly time series $\tau^1_1,\dots,\tau^{H}_t,\tau_2^1,\dots$ and not anymore as $H$-dimensional profiles.
For any day $t\in \{1,\dots,T\}$, and any half-hour $h \in \{1,\dots,H\}$ the smoothed temperature is defined by
\begin{equation}
\bar{\tau}^h_t = \left\{
    \begin{array}{ll}
        \tau^1_1 & \mbox{if } t=1 \mbox{ and } h=1 \\
         (1-a)\tau_t^h + a\bar{\tau}_{t}^{h-1} & \mbox{if } h \neq  1 \\
          (1-a)\tau_t^1 + a\bar{\tau}_{t-1}^{H} & \mbox{else,}\\
    \end{array}
\right.
\end{equation} 
where the smoothing parameter $a$ is in  $[0, 1]$. After testing several values, we set $a = 0.998$.
Then, for a given day $t\in \{1,\dots,T\}$, $\bar{\tau}_t$ is simply the daily average smoothed temperature: 
\begin{equation}
\bar{\tau}_t = \frac{1}{H} \sum_{h=1}^H \bar{\tau}^h_t\,.
\end{equation}

 \subsection{Causality Model} \label{app:causality_mod}
 
In the following, we detail models that were used to compute, for each household $i$ and for each tariff $p\in \mathcal{P}$, the daily profile of the mean and the standard deviation of household $i$ energy consumption.
We recall that these profiles are then used to cluster households.
Therefore, we have to estimate, for each half-hour $h$, the expectation and the standard deviation of the random variable $Y^{h}(i)\, | \, P=p$. 
To do so, we train a model that gives, for the tariff $p$ and the exogenous variables $x^h$, a forecast of the expected consumption at $h$ when tariff $p$ is selected and a forecast of the standard deviation of this consumption.
For any exogenous variable $x^h_t$ and tariff $p^h_t$, the random energy consumption $Y^{h}_t(i)$, of household $i$ at the half hour $h$ of the day $t$, is assumed Gaussian of mean $\mu_{i}(x^h_t,p^h_t)$ and standard deviation $\sigma_{i}(x^h_t,p^h_t)$. 
Moreover, we assume that these mean and standard deviation:
\begin{gather}
\mu^{i,h}(x^h_t, p^h_t)= \E\big[Y_t^h(i)\big] \quad
\mathrm{and} \quad \sigma^{i,h}(x^h_t, p^h_t)= \sqrt{\Var\big[Y_t^h(i)\big]},   
\end{gather}
 depend on additive smooth predictors.
Generalized additive models (GAM -- see~\cite{wood2006generalized}) may model electricity consumption (see~\cite{gaillard2016additive}) as a sum of independent exogenous variable effects.
Here, they are used to estimate conjointly, for any half-hour $h$ of a day $t$ and any tariff $p \in \mathcal{P}$, both $\mu^{i,h}(x^h_t,p)$ and $\sigma^{i,h}(x^h_t,p)$. These approximations are denoted by $\widehat{\mu}^{i},h(x^h_t,p)$ and $\widehat{\sigma}^{i,h}(p,x_t^h)$, respectively.
 For each half-hour $h$, we set the same underlying models: 
\begin{align}
\mu^{i,h}(x^h_t,p_t^h)= & s^{i,h}_{\tau}(\tau^h_t) + \xi_{\mathrm{L}}^{i,h} \mathbf{1}_{p_t^h=\mathrm{Low}} 
 \quad + \xi_{\mathrm{N}}^{i,h} \mathbf{1}_{p_t^h=\mathrm{Normal}}+ \xi_{\mathrm{H}}^h\mathbf{1}_{p_t^h=\mathrm{High}}  \\
\sigma^{i,h}(x^h_t,p_t^h) = &  \gamma_{\mathrm{L}}^{i,h} \mathbf{1}_{p_t^h=\mathrm{Low}}  
 \quad + \gamma_{\mathrm{N}}^{i,h} \mathbf{1}_{p_t^h=\mathrm{Normal}}+\gamma_{\mathrm{H}}^h\mathbf{1}_{p_t^h=\mathrm{High}} \,.
\label{eq:gam_elast}
\end{align}
where $s^{i,h}_{\tau}$, the function catching the effect of the temperature, is approximated by a cubic spline.
Fixing the number of knots $k$ and their positions is enough to determine a linear basis of dimension $k$ in which this function can be projected.
The \texttt{mgcv R-}package permits to estimate the coordinates of the spline in its basis and all the coefficients $\xi_{\mathrm{L}}^{i,h}$, $\xi_{\mathrm{N}}^{i,h}$, $\xi_{\mathrm{H}}^{i,h}$,  $\gamma_{\mathrm{L}}^{i,h}$, $\gamma_{\mathrm{N}}^{i,h}$, and $\gamma_{\mathrm{H}}^{i,h}$ defined in Equation~\eqref{eq:gam_elast}, which catch tariff effect. 
We highlight that models fitted on variances are linear.
 Both models (on mean and standard deviation) are estimated simultaneously, by setting the model family parameter of the \texttt{gam} function to the Gaussian location-scale model family.

Once the function and coefficients have been estimated (we write $\widehat{s}^{\,i,h}$ for the estimation of $s^{i,h}$ and so on),
for any tariff $p$, the estimations $\widehat{\mu}^{i,h}(x^h_t,p)$ and $\widehat{\sigma}^{i,h}(p,x_t^h)$ are computed:
\begin{align}
\widehat{\mu}^{i,h}(x^h_t,p)= &\widehat{s}^{\,i,h}_{\tau}(\tau^h_t) + \widehat{\xi}_{\mathrm{L}}^{i,h} \mathbf{1}_{p=\mathrm{Low}} + \widehat{\xi}_{\mathrm{N}}^{i,h} \mathbf{1}_{p=\mathrm{Normal}}+ \widehat{\xi}_{\mathrm{H}}^h\mathbf{1}_{p=\mathrm{High}}  \\ 
\mathrm{and}\quad \widehat{\sigma}^{i,h}(p,x_t^h) =  &
\widehat{\gamma}_{\mathrm{L}}^{i,h} \mathbf{1}_{p=\mathrm{Low}} +  \widehat{\gamma}_{\mathrm{N}}^{i,h} \mathbf{1}_{p=\mathrm{Normal}}+ \widehat{\gamma}_{\mathrm{H}}^h\mathbf{1}_{p=\mathrm{High}} \,.
\end{align}
Finally, the approximations of tariff impact are provided by: 
\begin{gather}
\E\big[\,Y^{i,h}\, | \, P=p\, \big] \approx \frac{1}{T}\sum_{t=1}^T \widehat{\mu}^{i,h}\big(p,x_t ^h\big)
\\
\mathrm{and} \quad \sqrt{\Var\big[\,Y^h\, | \, P=p\, \big]} \approx \frac{1}{T}\sum_{t=1}^T \widehat{\sigma}^{i,h}\big(p,x_t^h\big)\,.
\end{gather}

\subsection{Clustering method}
\label{app:clustering}

Here, the three steps of the clustering method proposed in Section~\ref{sec:clustering} are detailed.

\subsubsection{Scaling and gathering profiles} 

For an household $i \in \mathcal{I}$, for all $p\in \mathcal{P}$, the daily expected consumption profile $\mu^1_i(p),\dots, \mu^H_i(p)$ is considered. We assume that there is a base tariff $p_0 \in \mathcal{P}$ that corresponds to a signal of no incentive, namely Normal tariff.
We consider the quantity $\bar{\mu}_i=\frac{1}{H} \sum_{h=1}^H \mu^h_i(p_0)$  that is an approximation of the average daily expected consumption of household $i$ 
under no DR program. \\

Then, all the profiles of household $i$ are rescaled by this quantity and, for each tariff $p \in \mathcal{P}$, the daily consumption profiles under tariff $p$ of all the households $i \in \mathcal{I}$ are gathered in a matrix $\M(p) \in \mathcal{M}_{|\mathcal{I}|\times H}$.
Finally, the matrix $\M \in \mathcal{M}_{|\mathcal{I}|\times H|\mathcal{P}|}$ is created by binding by column matrices $\M(p)$, so 
\begin{gather}
\M_{i,h} (p) = \frac{\mu_i^h(p)}{\bar{\mu}_i}  \quad  \mathrm{and} \quad \M=\bigg(
\M\big(1\big)  \, \, \Big|  \,\, \dots   \,\, \Big| \, \, \M\big(P\big) \bigg)\,.
\end{gather}

\paragraph{Low Rank Approximation.}
Since we are interested in energy consumption, all the coefficients of $\M$ are non-negative~-- we will write $\M \geq 0$ and say that this matrix is non-negative.
To reduce dimension of non-negative matrices, the factorization method proposed by~\cite{paatero1994positive} and \cite{lee1999learning} that uses non-negativity constraints is considered.
The integer $r \ll \min(|\mathcal{I}|,H|\mathcal{P}|)$ that will ensure a reduction of the dimension is fixed (we chose $r=5$ in our case study). 
The non-negative matrix factorization (NMF) approximates matrix of profiles $\M$ with $\W\H$ by minimizing the euclidean distance between both matrices under the constraint that $\W$ and $\H$ are non-negative matrices of size $|\mathcal{I}| \times r$ and $r \times H|\mathcal{P}|$, respectively.
Function \texttt{NMF} of the \texttt{Python}-library \texttt{sklearn.decomposition} allows to approximate $\W$ and $\H$ with a coordinate descent solver. 
For simplicity of notation, $\W$ is confounded with its approximation.
Thus, for any household $i \in \mathcal{I}$, we get $r$ features, namely the $i^{\mathrm{th}}$ line of matrix $\W$, that we denote by $\W_{i \, \cdot}$ in the following.

\subsubsection{$k$-medoid clustering}

Now, the vectors $\W_{i \, \cdot}$ allow to cluster households in $k$ clusters.
In $k$-means clustering, the center of a given cluster is simply the average between the points of this cluster. Since it can be influenced by extreme value, $k$-means algorithm is sensitive to outliers. Conversely, $k$-medoid algorithm chooses data points to represent clusters, which makes it more robust and favors a clustering where clusters have sizes of the same order. This algorithm was introduce by \cite{kaufman1987clustering} with the $L^1$-norm. Here, we use it with the Euclidean distance and the best clustering $C^{\star}_1,\dots,C^{\star}_k$ is the one that minimizes the following criteria: 
\begin{gather}
\Big\{C^{\star}_1,\dots,C^{\star}_k\Big\} \in \argmin_{\{C_1,\dots,C_k\}} 
\sum_{\ell=1}^k \sum_{i \in C_{\ell}} \big\| \W_{i \, \cdot}- \W_{C_\ell \, \cdot}  \big\|^2\, \quad
\mathrm{with} \quad C_\ell \in \argmin_{i \in C_{\ell}} \sum_{j \in C_{\ell}} \big\| \W_{i \, \cdot}- \W_{j \, \cdot}  \big\|^2\,,
\end{gather}
where $\|\cdot\|$ is the Euclidean norm.
The clusters are computed by using \texttt{KMedoid} function implemented in the  \texttt{Python}-library \texttt{sklearn\_extra}.

 \subsection{Variational Autoencoder} \label{app:vae}
 
The calculations below are an adaptation of the ones proposed by~\cite{kingma2013auto} to our case-study.
The generation of the data is assumed to follow a two-steps process: firstly, a variable $Z$ was sampled from a standard Gaussian and then, $Y$ was sampled from the distribution $p_{\theta^{\star}}( \, \cdot \, | Z)$. 
 The decoder, parametrized by $\theta$, can model this process: with $Z \sim \mathcal{N}\big(0,I_{d}\big)$ as  input, it generates the variable $Y$, conditionally to $Z$, by sampling it from $p_{\theta}( \, \cdot \, | Z)$, which is an approximation of the true distribution $p_{\theta^{\star}}( \, \cdot \, | Z)$.
In our generation process, we will denote by $q_{_Y}(Z)$ the approximation made by the encoder of the density of $Z\, |\, Y$.
The variational autoencoder is trained in a way that $q_{_Y}$ is the Gaussian of mean $\mu(Y)$ and covariance matrix $\Sigma(Y)$, where $\mu(Y)$ and $\Sigma(Y)$ are the outputs of the encoder for the input $Y$.
For $Y \in \R^H$, by using Bayes' theorem and the variables $Z$ sampled from the encoder distribution $q_{_Y}$, the log-marginal likelihood $p_{\theta}(Y) $ satisfies
\begin{align}
\log p_{\theta}(Y) & = \, \E_{Z \sim q_{Y}} \big[ \log p_{\theta}(Y) \big] 
= \, \E_{Z \sim q_{Y}} \bigg[ \log \frac{ p_{\theta}(Y|Z) \, p_{\theta}(Z)}{p_{\theta}(Z|Y)} \bigg]  \notag \\
= \, & \E_{Z \sim q_{Y}}  \bigg[ \log \frac{q_{Y}(Z)}{p_{\theta}(Z |Y)}   +  \log \frac{p_{\theta}(Z)}{ q_{Y}(Z)}   +\log {p_{\theta}(Y|Z)}   \bigg] \notag\\
 = \, & \kl \big(q_{Y} (Z) \, || \, p_{\theta}(Z|\,Y) \big) - \kl \big( q_{Y} (Z)\,  ||  \,p_{\theta}(Z) \big) + \, \E_{Z \sim q_{Y}} \big[  \log {p_{\theta}(Y|\,Z)} \big] \,. 
\end{align}
The first term corresponds to the error made by approximating the distribution $\p_{\theta}(\,\cdot\,|\,Y)$ with $q_{_Y}$. Thus to conjointly maximizing the log-likelihood and minimizing this approximation error, the loss
\begin{equation}
 \kl \big( q_{_Y}(Z) \,  ||  \, p(Z) \big) - \E_{Z \sim q_{_Y}} \big[  \log {p_{\theta}(Y |Z)} \big] \,,
 \label{eq:kl_logv}
\end{equation}
 has to be minimized. The two parts of the equation above are known as the regularization term and the reconstruction term, respectively. 
We recall that $q_{_Y}$ is the Gaussian distribution of mean $\mu(Y)$ and of covariance matrix $\Sigma(Y)$ and that we assume $Z \sim \mathcal{N}\big(0,I_{d}\big)$, so the regularization term is the Kullback–Leibler divergence between $\mathcal{N}\big(\mu(Y),\Sigma(Y)\big)$ and a standard $d$-multidimensional normal distribution. 
Moreover, we highlight that if the decoder samples $Y | Z$ from a distribution
of the exponential family,
\begin{equation}
p_{\theta}(Y|Z)  = a(Y)b(Z)\exp \big(\eta(Z)T(Y)\big) \,,
\end{equation}
with $\theta$ gathering the functions  $a$, $b$, $\eta$, and $T$. Then, the second term is explicit.  
But, for a given $Z$, the decoder outputs a unique vector $\mathrm{D}(Z)=\widehat{Y}$, so inferring the previous distribution is a tough task. 
Nevertheless, assuming that $Y \, |\, Z$ is a multivariate Gaussian of mean $\mathrm{D}(Z)$ and with a known covariance matrix $\sigma^2I_{d}$, a very simple expression of the regularization term is obtained:
\begin{equation} 
- \log p_{\theta}(Y|Z) = \frac{1}{2\sigma^2} \big\| Y-\mathrm{D}(Z)\big\|_2^2 - \log \big( 2\pi^{H/2} \sigma \big) \,.
\end{equation}
Therefore, the loss defined in Equation~\eqref{eq:kl_logv} can be re-written:
\begin{equation} 
\frac{1}{2\sigma^2} \big\| Y- \widehat{Y} \big\|_2^2  + \kl \big( \mathcal{N}\big(\mu(Y),\Sigma(Y)\big) \,  ||  \, \mathcal{N}(0,I_{d}) \big) \,. 
\end{equation}
Under all the assumptions above, and given the independent observations $Y_1,\dots Y_{T_0}$, to obtain the generative process that  best models the real one, we will thus consider the loss 
 \begin{align}
L_{\mathrm{VAE}}&(\eta) = \frac{1}{T_0}\sum_{t=1}^{T_0} \Bigg( \, \big\| Y_t-\widehat{Y}_t \big\|^2     +  \eta \, \kl \Big( \mathcal{N}\big(\mu(Y_i),\Sigma(Y_i)\big) \,  \big|\big|  \, \mathcal{N}(0,I_{d}) \Big) \Bigg)\,.
\end{align}
We recall that the vectors $\widehat{Y}_t$ are the outputs of the decoder $\mathrm{D}(Z_t)$, where the random variable is sampled from a $d$-multivariate Gaussian of mean $\mu(Y_t)$ and covariance matrix $\Sigma(Y_t)$.
This loss is conjointly maximizing the log-likelihood of the observation with the data generation process distribution:
\begin{equation}
\log p_{\theta} (Y_1,\dots Y_t) = \sum_{t=1}^{T_0}  \log p_{\theta} (Y_t)\,
 \end{equation}
 and minimizing the approximation error 
 \begin{equation}
 \sum_{t=1}^{T_0} \kl \big(q_{Y_t}  (Z) \, || \, p_{\theta}( Z |\, Y_t) \big)\,.
 \end{equation}
It is important to underline that the previous calculations are still valid when  all the distributions are conditioned by exogenous variables.

\subsection{Correlation matrix of semi-parametric generator} \label{app:semi_param}

Here the estimation of the matrix $\Sigma$, which is used to generate profiles with correlations between temporal intervals of the same day, is detailed.
If the model defined by Equation~\eqref{eq:semi_param_mod} was true, residuals $Y_t^h-f^h(x_t^h,p_t^h)$ should be Gaussian of mean $0$ and standard deviation $\sigma^h(p_t^h)$.
Thus the vector of standardized residuals $\e_t = (e^h_t)_{1 \leq h\leq H}$ is considered, where
\begin{equation}
e^h_t=\frac{Y_t^h-f^h(x_t^h,p_t^h)}{\sigma^h(p_t^h)}\,.
\end{equation}
Assuming the model above, the covariance matrix $\Sigma$ of vectors $\e_1, \dots, \e_{T_0}$ should have $1$ on the diagonals and all other coefficients between $-1$ and $1$.
To deal with our imperfect modeling and avoid again the problem of high standard deviation coming from the estimation error, $\Sigma$ is approximated by the empirical correlation matrix of vectors $\e_1, \dots, \e_{T_0}$.
From the ${T_0}$ observations $\e_1,\dots,\e_{T_0}$, which are assumed independent, of the $H$-dimensional random vector $\e=(e^1,\dots,e^H)$, the coefficients of the $H\times H$-correlation matrix $\Sigma$ are defined by
\begin{gather}
\Sigma_{i,j} = \frac{\mathrm{cov}(e^i,e^j)}{\sqrt{\Var(e^i) \Var(e^j)}}\,, \quad
\mathrm{where} \quad  \mathrm{cov}(e^i,e^j) = \E(e^i e^j) - \E(e^i) \E(e^j)\,.
\end{gather}
We point out that in the case of random variables $\e^1,\dots, \e^H$ of standard deviation $1$ (we assume it in the semi-parametric simulator described in Section~\ref{sec:semi_param}), covariance and correlation matrices are equal. 
In there, $\Sigma_{i,j}$ is estimated by replacing covariances and variances of random variables $e^i$ and $e^j$ by the their empirical estimations:
\begin{align}
\mathrm{cov}(e^i,e^j) \approx \frac{1}{T_0-1} \sum_{t=1}^{T_0} & \big(e^i_t - \bar{e}^{\,i}\big) \big(e^j_t - \bar{e}^{\,j}\big) \quad
\mathrm{and} \quad 
\Var(e^i)  \approx \frac{1}{T_0-1} \sum_{t=1}^{T_0}& \big(e^i_t - \bar{e}^{\,i}\big)^2  \,, \quad
 \mathrm{with} \quad 
\bar{e}^{\,i}=  \frac{1}{T_0} \sum_{t=1}^{T_0} e^i_t\,.\notag
\end{align}

\subsection{Experiments - graphical results}
\label{app:graphs}
\begin{figure*}[hb!]
\centering
\includegraphics[width=0.47\textwidth]{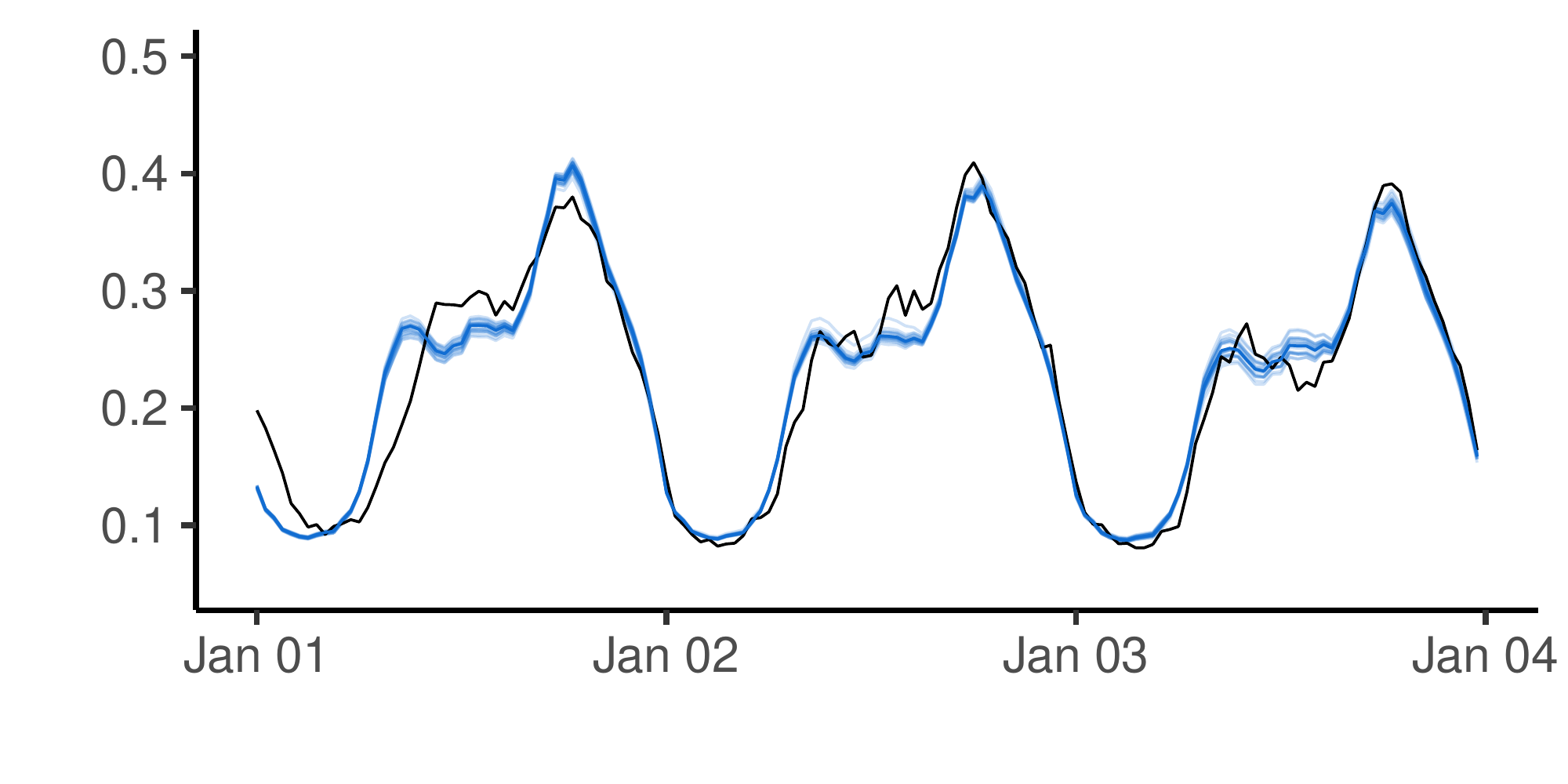}  \hfill
\includegraphics[width=0.52\textwidth]{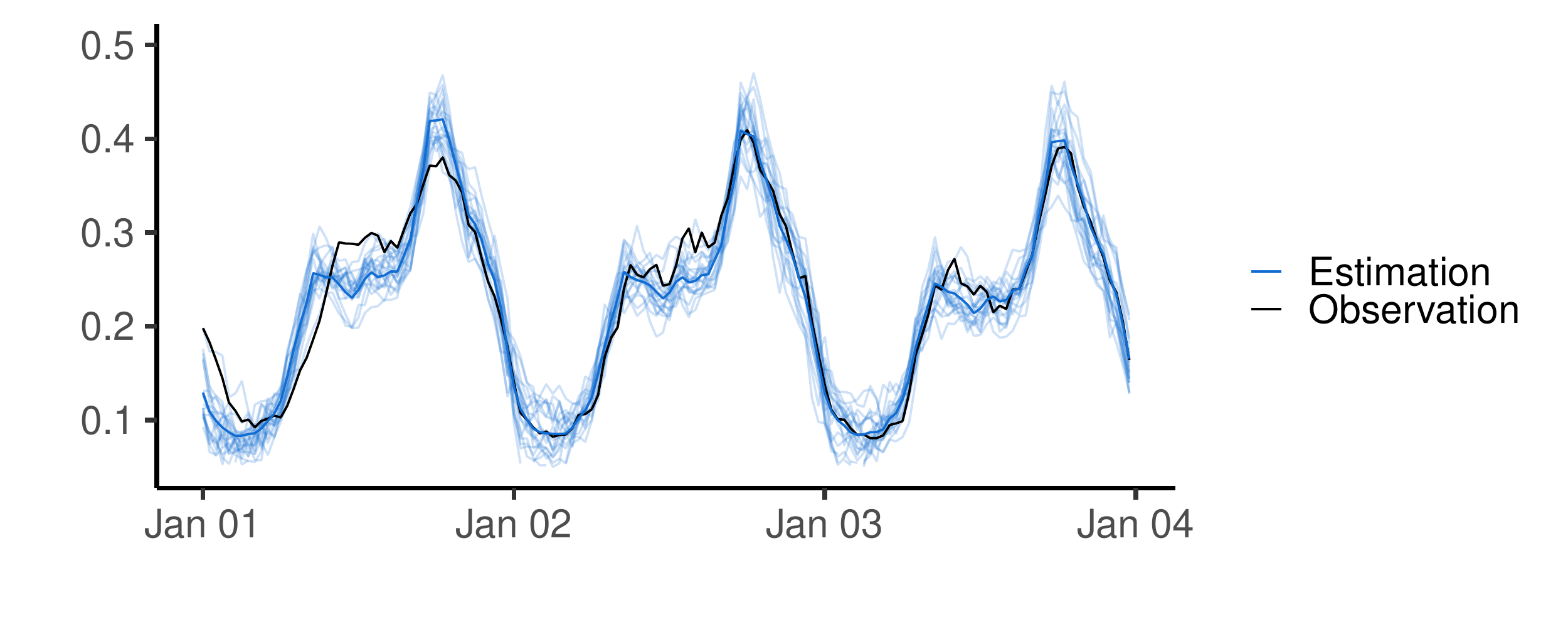}
 \includegraphics[width=0.47\textwidth]{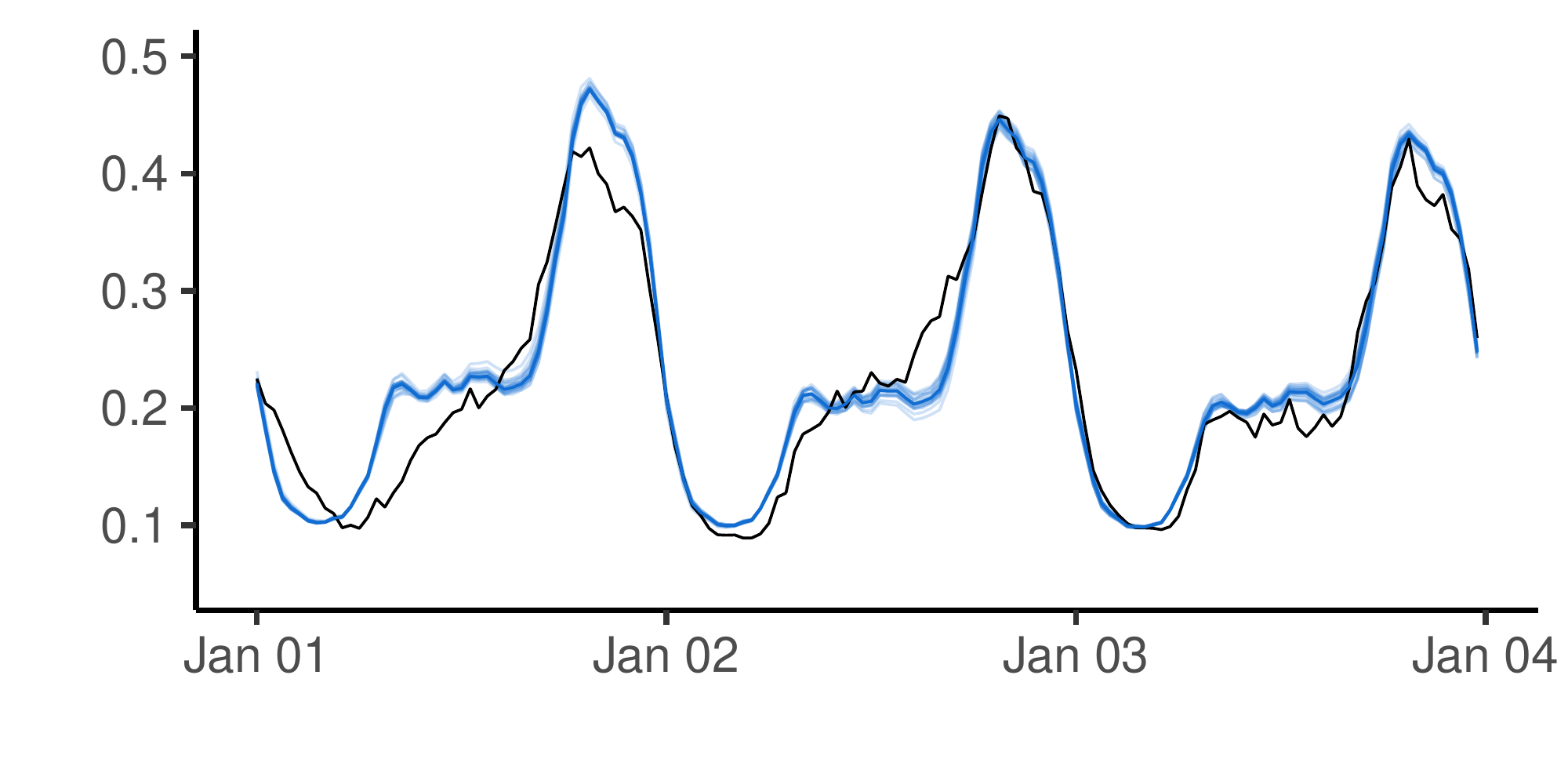}  \hfill
\includegraphics[width=0.52\textwidth]{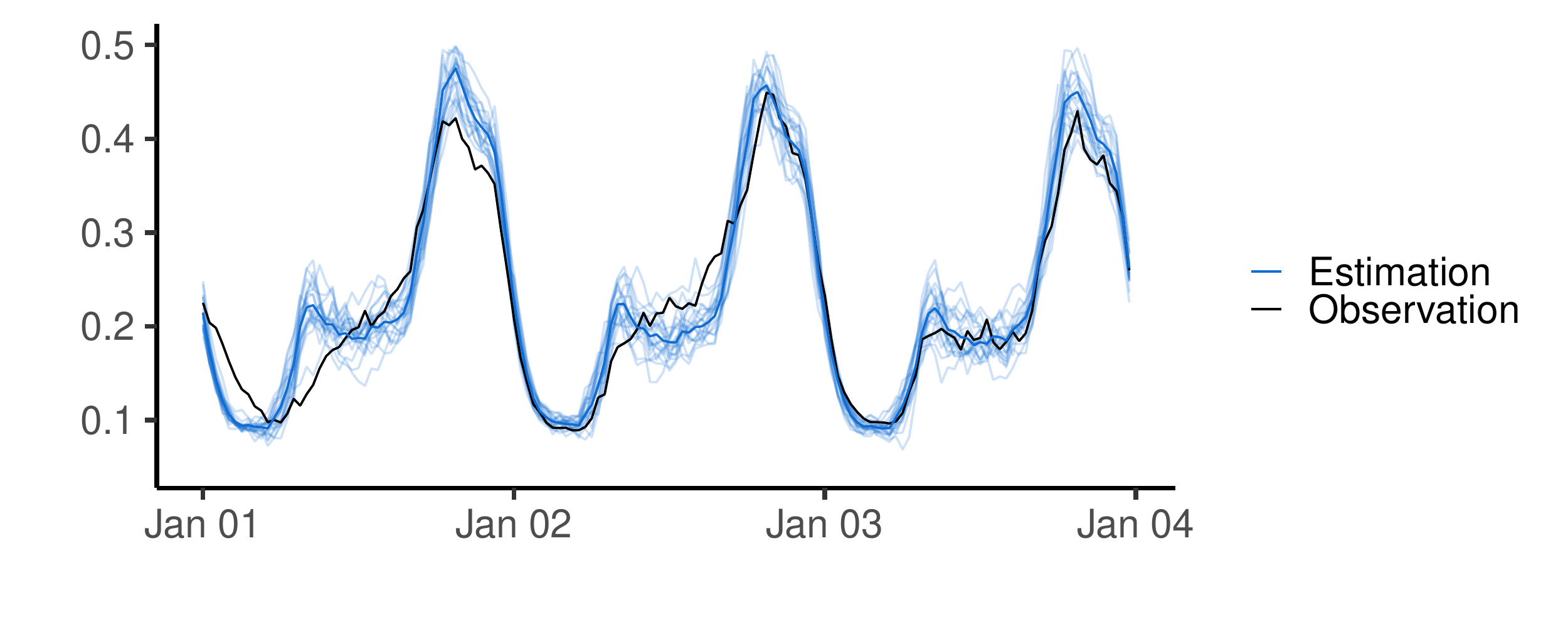} 
\includegraphics[width=0.47\textwidth]{estimations_cluster_2_cvae.pdf}  \hfill
\includegraphics[width=0.52\textwidth]{estimations_cluster_2_gam.pdf} 
\includegraphics[width=0.47\textwidth]{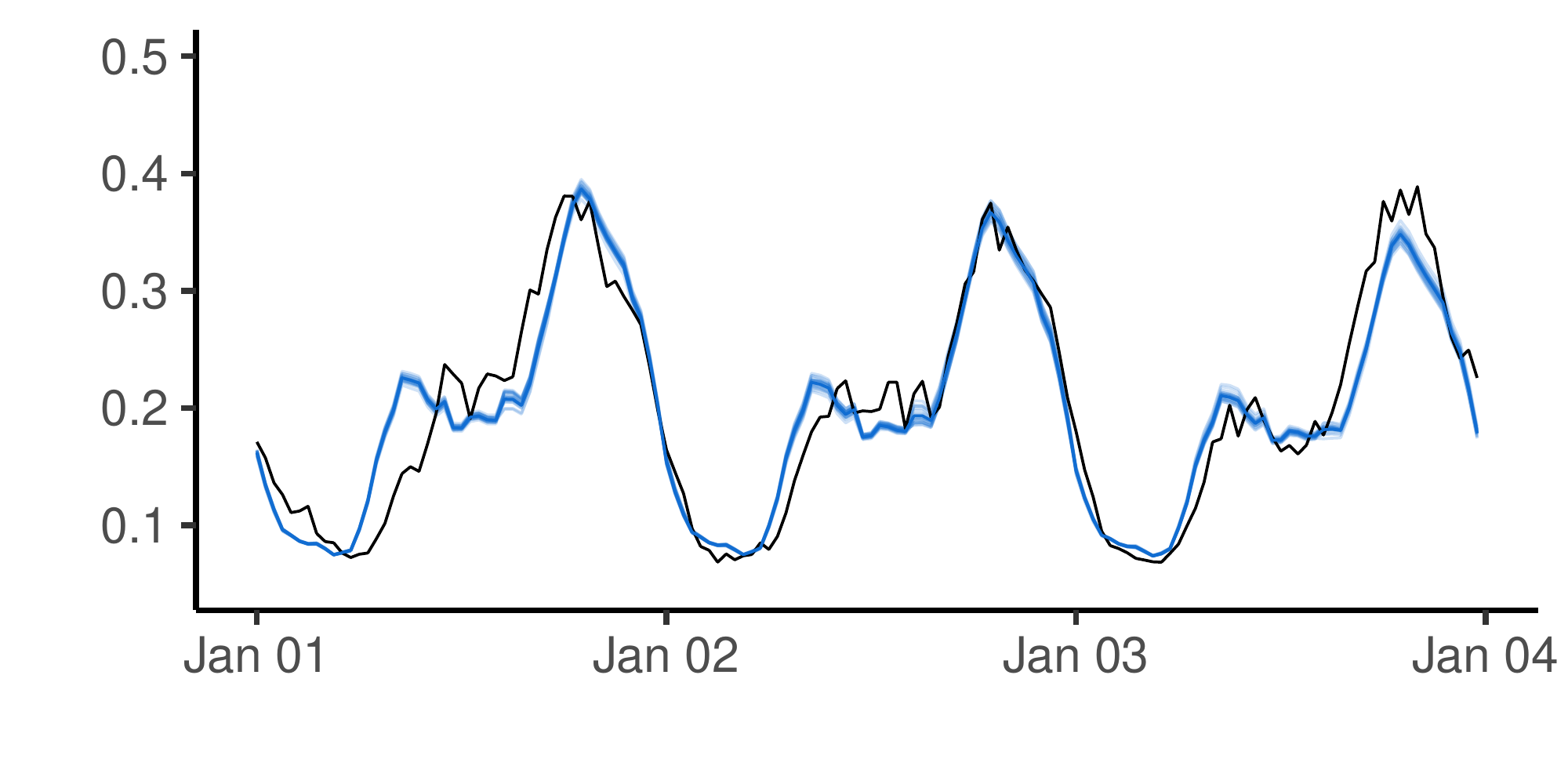}  \hfill
\includegraphics[width=0.52\textwidth]{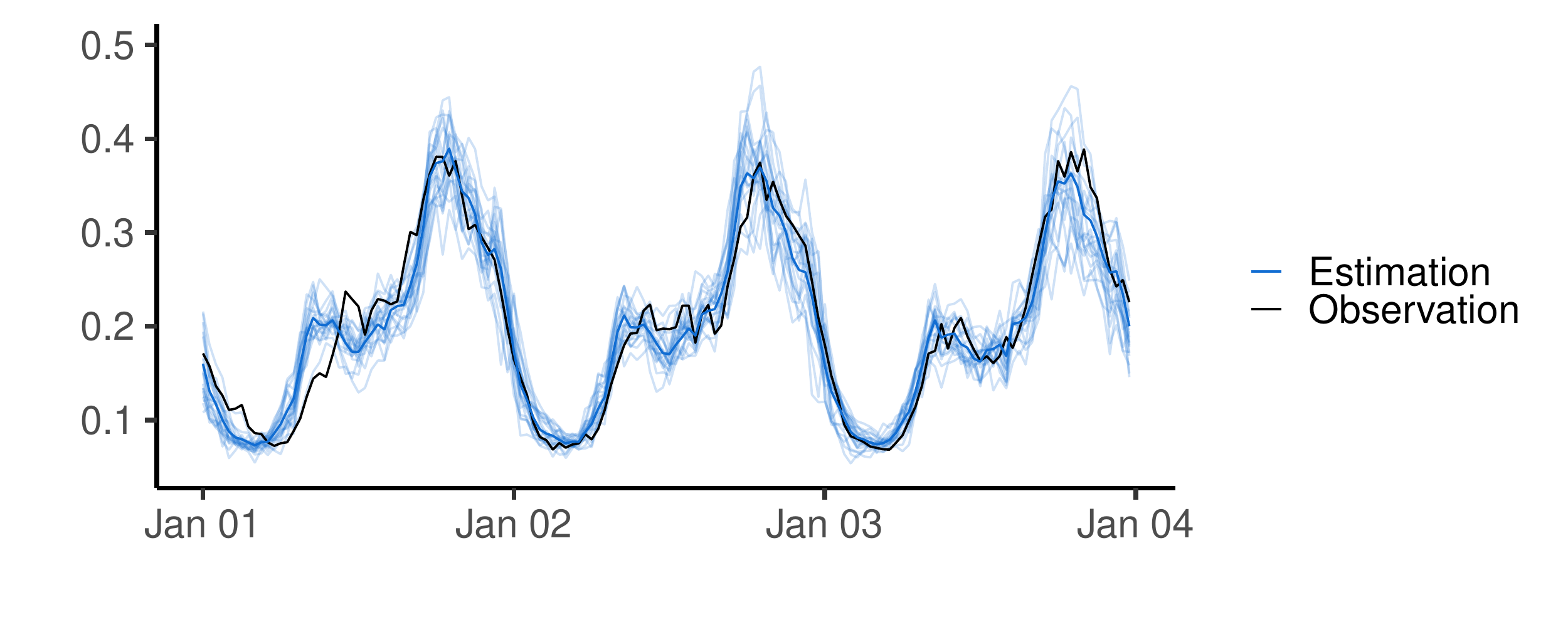} 
\caption{Left: data generated with the CVAE-based generator.  Right: data generated with the GAM-based generator. 
Blue lines: for every cluster over the first three days of the testing set, $20$ energy consumption profiles and empirical mean profile, calculated on 2 $200$ samples (in bold), obtained by giving, to the two simulators, the exogenous variables observed over this period.
Black line: real observed profiles.}
\label{fig:estimations_all}
\end{figure*}

\begin{figure*}[h!]
\centering
\includegraphics[width=0.47\textwidth]{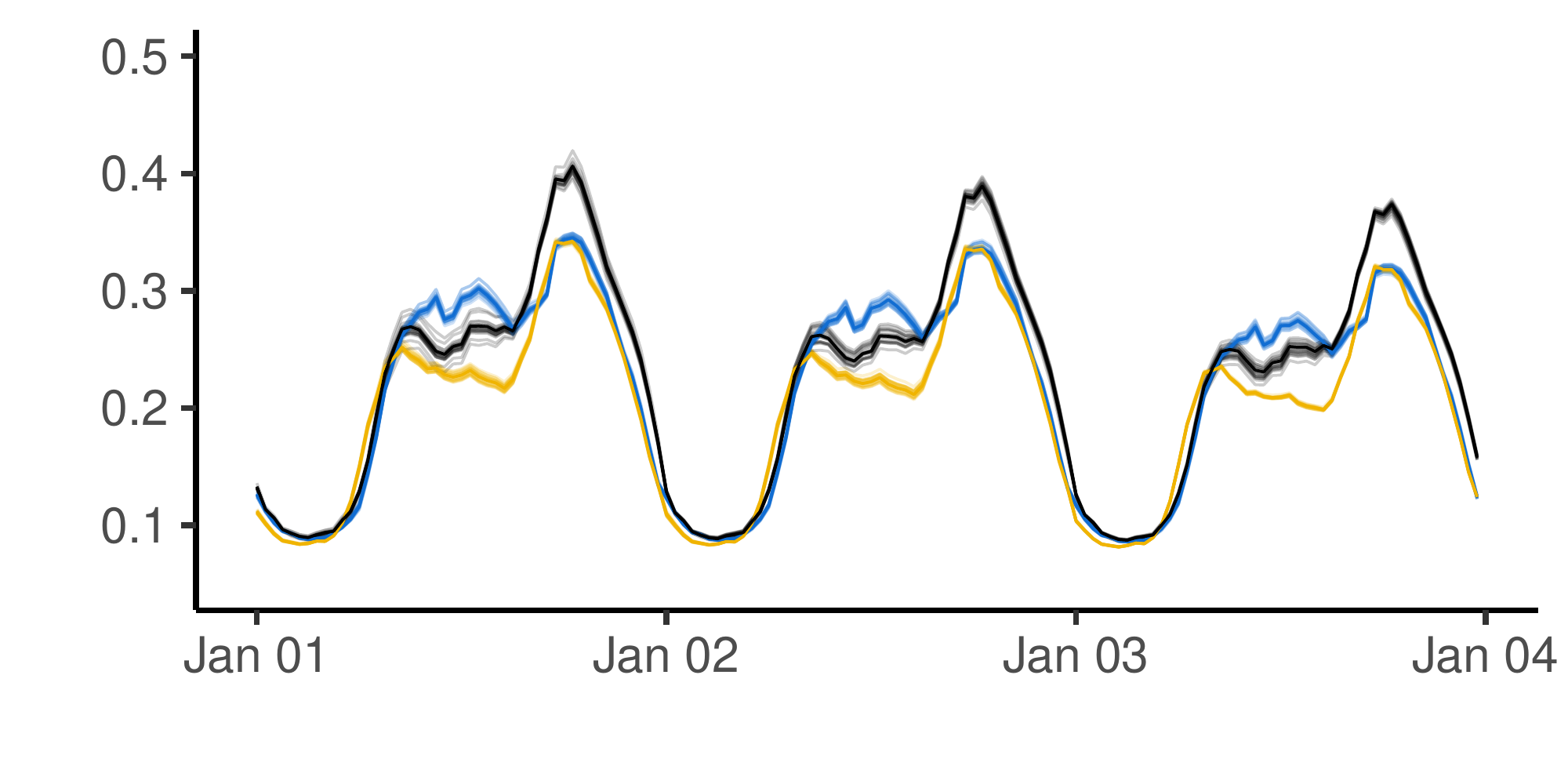}  \hfill
\includegraphics[width=0.52\textwidth]{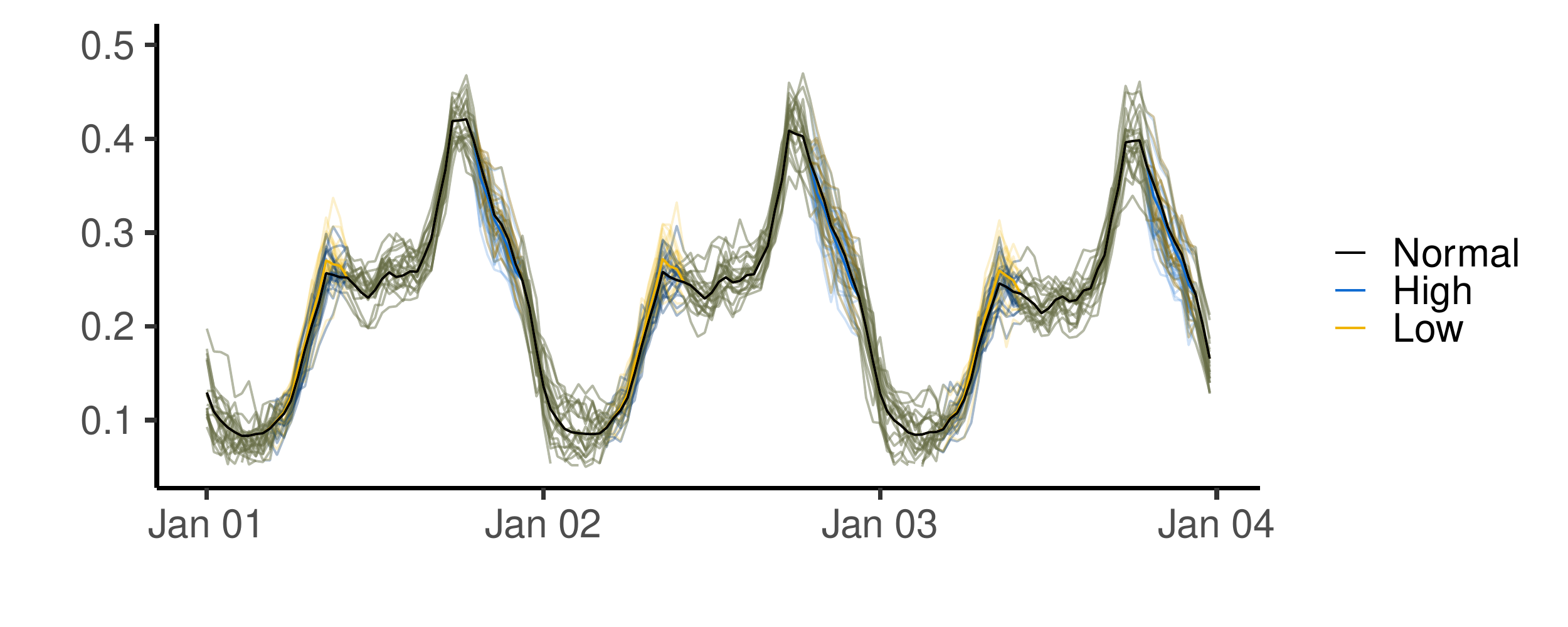} 
\includegraphics[width=0.47\textwidth]{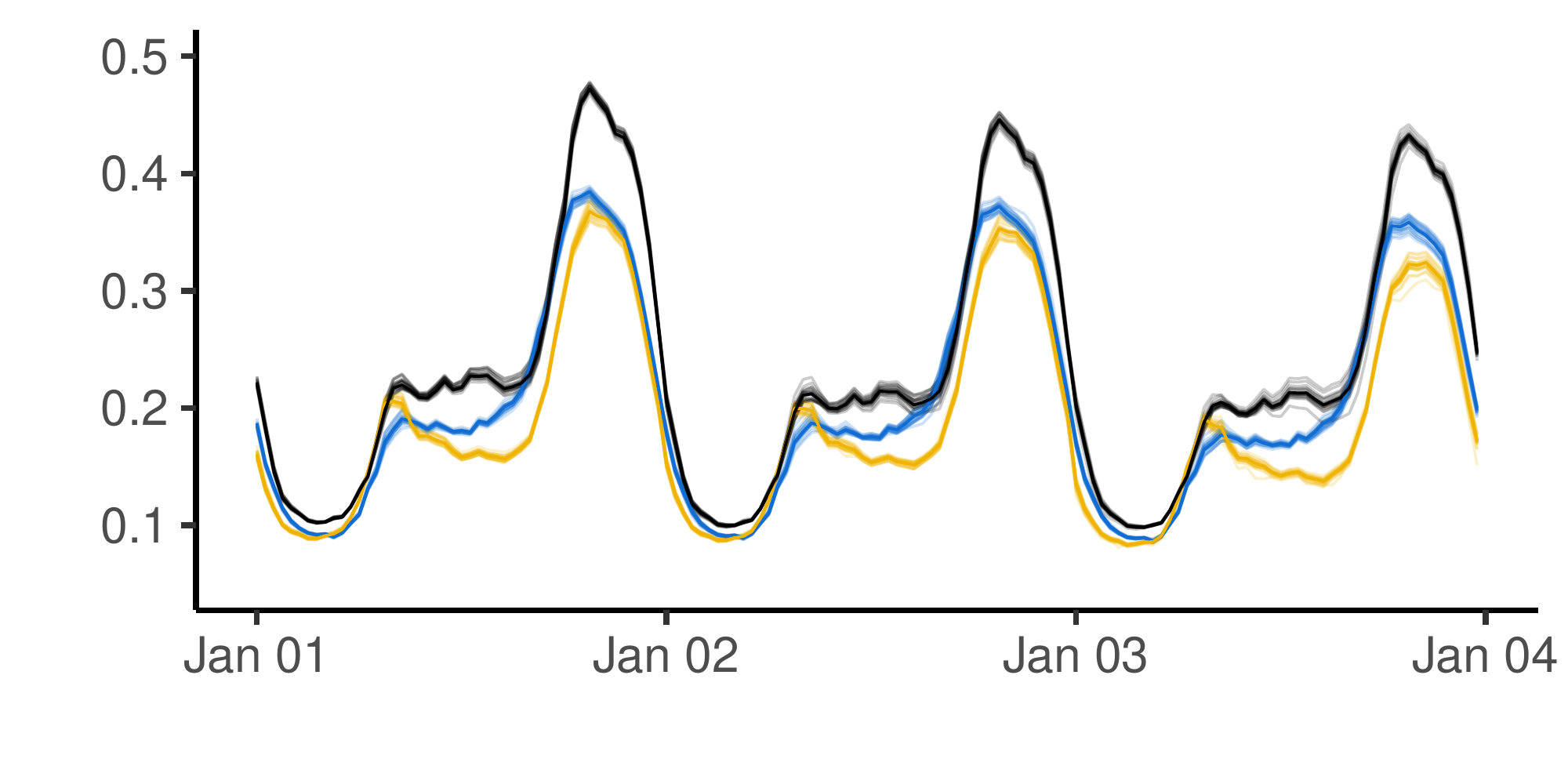}  \hfill
\includegraphics[width=0.52\textwidth]{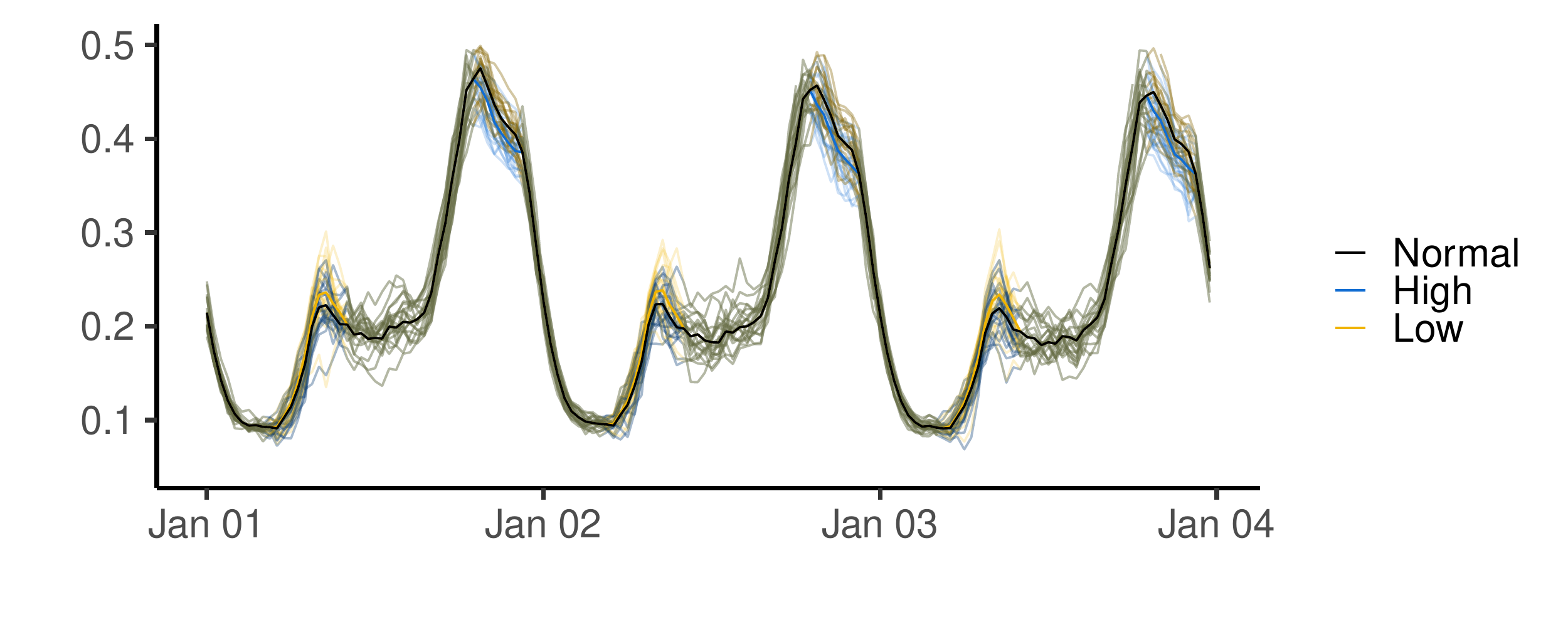} 
\includegraphics[width=0.47\textwidth]{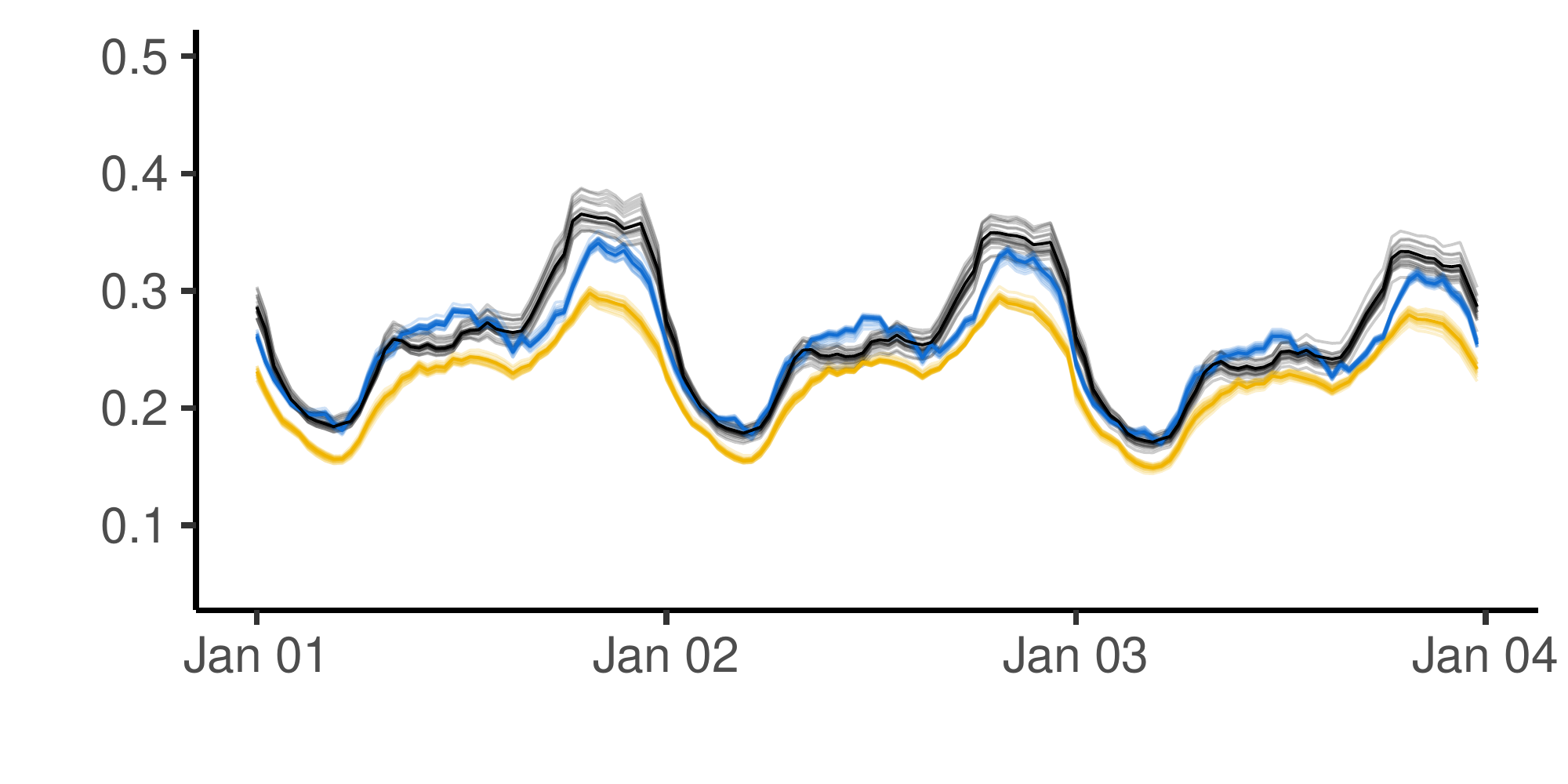}  \hfill
\includegraphics[width=0.52\textwidth]{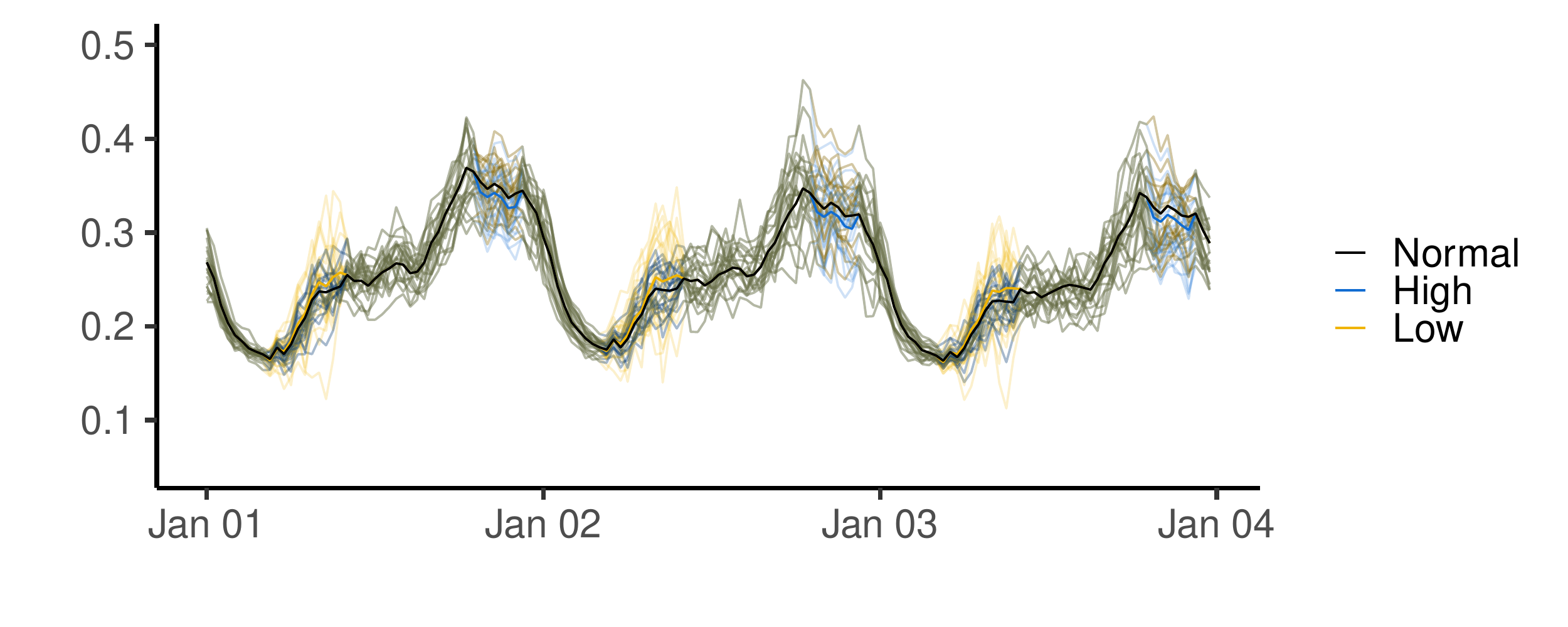} 
\includegraphics[width=0.47\textwidth]{simulations_cluster_3_cvae.pdf}  \hfill
\includegraphics[width=0.52\textwidth]{simulations_cluster_3_gam.pdf} 
\caption{Left: data generated with the CVAE-based generator.  Right: data generated with the GAM-based generator.
Black lines: for every cluster on the first three days of the testing set, $20$ energy consumption profiles and empirical mean profile, computed over $200$ samples (in bold), obtained by giving, to the two simulators, a Normal tariff for every half-hour and the weather and calendar variables observed over this period.
Blue lines: same plots but with a High tariff in the evening and Normal tariff otherwise.
Yellow lines: same plots but with a Low tariff in the early morning and Normal tariff otherwise.}
\label{fig:simulations_all}
\end{figure*}

\end{document}